# UNetFormer: A UNet-like Transformer for Efficient Semantic Segmentation of Remote Sensing Urban Scene Imagery

Libo Wang[1, 2], Rui Li[1], Ce Zhang[3, 4], Shenghui Fang[1*], Chenxi Duan[5], Xiaoliang Meng[1, 2] and Peter M. Atkinson[3, 6, 7]

1) School of Remote Sensing and Information Engineering, Wuhan University, 129 Luoyu Road, Wuhan, Hubei 430079, China.

2) Key Laboratory of Natural Resources Monitoring in Tropical and Subtropical Area of South China, Ministry of Natural Resources, Guangzhou, Guangdong 510000, China.

3) Lancaster Environment Centre, Lancaster University, Lancaster LA1 4YQ, UK.

4) UK Centre for Ecology & Hydrology, Library Avenue, Lancaster LA1 4AP, UK.

5) Faculty of Geo-Information Science and Earth Observation, University of Twente, Enschede, the Netherlands.

6) Geography and Environmental Science, University of Southampton, Highfield, Southampton SO17 1BJ, UK.

7) Institute of Geographic Sciences and Natural Resources Research, Chinese Academy of Sciences, 11A Datun Road, Beijing 100101, China.

*Corresponding author.

*Abstract*—Semantic segmentation of remotely sensed urban scene images is required in a wide range of practical applications, such as land cover mapping, urban change detection, environmental protection, and economic assessment. Driven by rapid developments in deep learning technologies, the convolutional neural network (CNN) has dominated semantic segmentation for many years. CNN adopts hierarchical feature representation, demonstrating strong capabilities for local information extraction. However, the local property of the convolution layer limits the network from capturing the global context. Recently, as a hot topic in the domain of computer vision, Transformer has demonstrated its great potential in global information modelling, boosting many vision-related tasks such as image classification, object detection, and particularly semantic segmentation. In this paper, we propose a Transformer-based decoder and construct an UNet-like Transformer (UNetFormer) for real-time urban scene segmentation. For efficient segmentation, the UNetFormer selects the lightweight ResNet18 as the encoder and develops an efficient global-local attention mechanism to model both global and local information in the decoder. Extensive experiments reveal that our method not only runs faster but also produces higher accuracy compared with state-of-the-art lightweight models. Specifically, the proposed UNetFormer achieved 67.8% and 52.4% mIoU on the UAVid and LoveDA datasets, respectively, while the inference speed can achieve up to 322.4 FPS with a 512×512 input on a single NVIDIA GTX 3090 GPU. In further exploration, the proposed Transformer-based decoder combined with a Swin Transformer encoder also achieves the state-of-the-art result (91.3% F1 and 84.1% mIoU) on the Vaihingen dataset. The source code will be freely available at https://github.com/WangLibo1995/GeoSeg.

# 1. Introduction

Driven by advances in sensor technology, fine-resolution remotely sensed urban scene images have been captured increasingly across the globe, with abundant spatial details and rich potential semantic contents. Urban scene images have been subjected extensively to semantic segmentation, the task of pixel-level segmentation and classification, leading to various urban-related applications, including land cover mapping (Li et al., 2022b; Maggiori et al., 2016; Marcos et al., 2018), change detection (Xing et al., 2018; Yin et al., 2018), environmental protection (Samie et al., 2020), road and building extraction (Griffiths and Boehm, 2019; Shamsolmoali et al., 2020; Vakalopoulou et al., 2015) and many other practical applications (Picoli et al., 2018; Shen et al., 2019). Recently, a growing wave of deep learning technology (LeCun et al., 2015), in particular the convolutional neural network (CNN), has dominated the task of semantic segmentation (Chen et al., 2014; Chen et al., 2018b; Long et al., 2015; Ronneberger et al., 2015; Zhao et al., 2017a). Compared with traditional machine learning methods for segmentation, such as the support vector machine (SVM) (Guo et al., 2018), random forest (Pal, 2005) and conditional random field (CRF) (Krähenbühl and Koltun, 2011), CNN-based methods are capable of capturing more fine-grained local context information, which underpins its huge capabilities in feature representation and pattern recognition (Zhang et al., 2020a; Zhang et al., 2020b).

Despite the above advantages, the convolution operation with a fixed receptive view is designed to extract local patterns and lacks the ability to model global contextual information or long-range dependencies in its nature. As for semantic segmentation, per-pixel classification is often ambiguous if only local information is modelled, while the semantic content of each pixel becomes more accurate with the help of global contextual information (Yang et al., 2021a) (Li et al., 2021c). The global and local contextual information is illustrated in **Fig. 1**. Although the self-attention mechanism alleviates the above issue (Vaswani et al., 2017) (Wang et al., 2018), they normally require significant computational time and memory to capture the global context, thus, reducing their efficiency and restricting their potential for real-time urban applications.

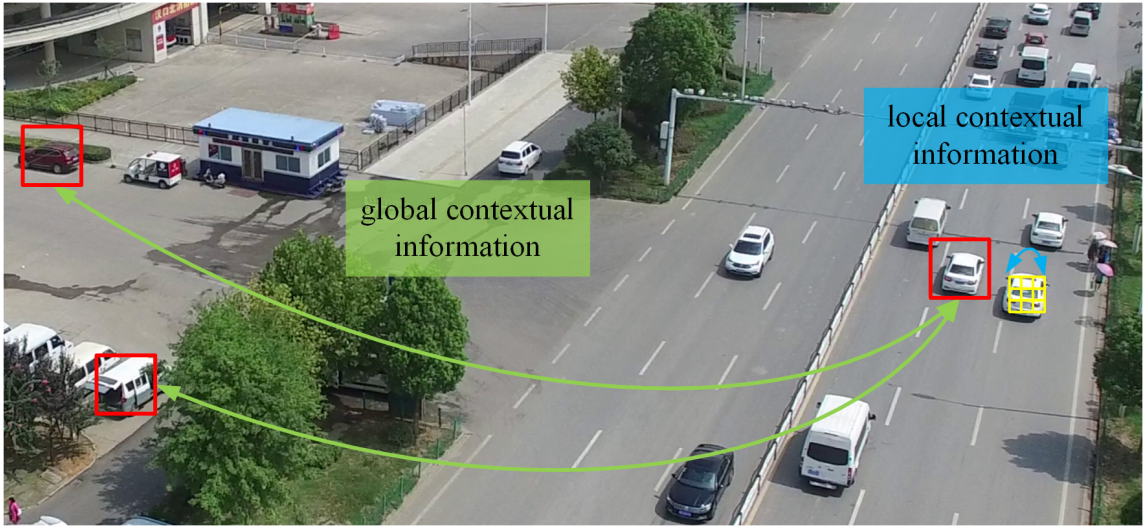

**Fig. 1** Illustration of the global and local contextual information. The local contextual information is modelled by convolutions (yellow). The global contextual information is modelled by long-range window-wise dependencies (red).

In this paper, we aim to achieve precise urban scene segmentation while ensuring the efficiency of the network simultaneously. Inspired by the recent breakthrough of Transformers in computer vision, we propose a **UNet**-like Trans**former** (UNetFormer) to address such a challenge. The UNetFormer innovatively adopts a hybrid architecture consisting of a CNN-based encoder and a specifically designed Transformer-based decoder. Specifically, we adopt the ResNet18 as the encoder and design a global-local Transformer block (GLTB) to construct the decoder. Unlike the conventional self-attention block in the standard Transformer, the proposed GLTB develops an efficient global-local attention mechanism with an attentional global branch and a convolutional local branch to capture both global and local contexts for visual perception, as illustrated in **Fig. 2**. In the global branch, the window-based multi-head self-attention and cross-shaped window context interaction module are introduced to capture global contexts with low complexity (Liu et al., 2021). In the global branch, convolutional layers are applied to extract the local context. Finally, to effectively fuse the spatial details and context information as well as further refine the feature maps, a feature refinement head (FRH) is proposed and attached at the end of the network. The trade-off between accuracy and efficiency as well as effective feature refinement allows the proposed method to exceed the state-of-the-art lightweight networks for efficient segmentation of remotely sensed urban scene images, demonstrated by four public datasets: the UAVid (Lyu et al., 2020), ISPRS Vaihingen and Potsdam datasets, as well as the LoveDA (Wang et al., 2021a).

The remainder of this paper is organized as follows. In Section 2, we review the related work on CNN-based and Transformer-based urban scene segmentation and global context modelling. In Section 3, we present the structure of our UNetFormer and introduce the proposed GLTB and FRH. In Section 4, we conduct an ablation study to demonstrate the effectiveness of GLTB and FRH as well as the novel hybrid structure and compare the results with a set of state-of-the-art models applied to the four datasets. In Section 5, we provide a comprehensive discussion. Section 6 is a summary and conclusion.

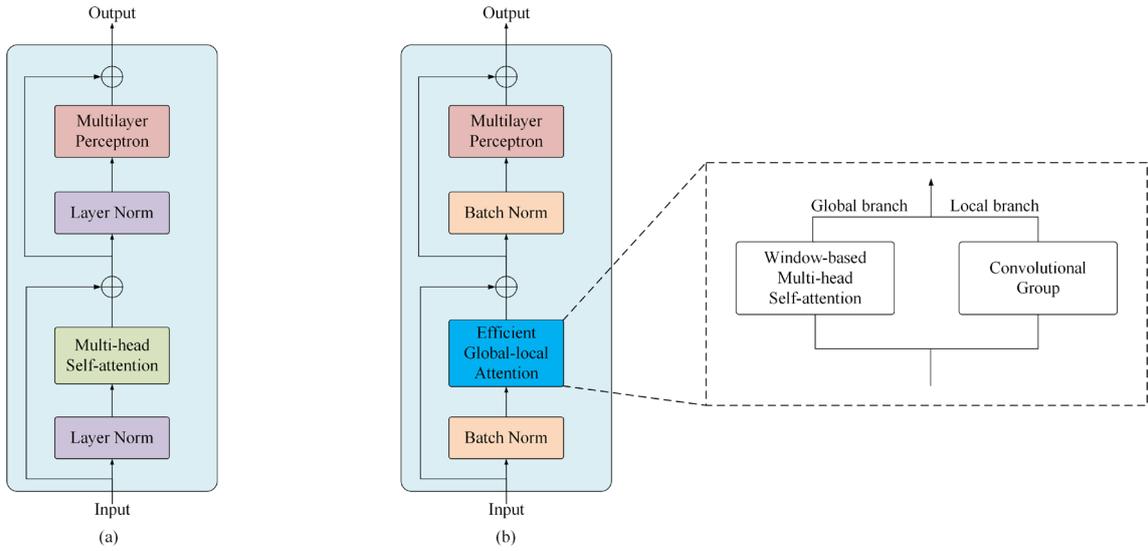

**Fig. 2** Illustration of (a) the standard Transformer block and (b) the global-local Transformer block.

## 2. Related work

### 2.1 CNN-based semantic segmentation methods

The fully convolutional network (FCN) (Long et al., 2015) is the first effective CNN structure to address semantic segmentation problems in an end-to-end manner. Since then, CNN-based methods have dominated the semantic segmentation task in the remote sensing field (Kemker et al., 2018; Kotaridis and Lazaridou, 2021; Ma et al., 2019; Tong et al., 2020; Zhao and Du, 2016; Zhu et al., 2017). However, the over-simplified decoder of FCN leads to a coarse-resolution segmentation, limiting the fidelity and accuracy.

To address this problem, an encoder-decoder network, i.e., the UNet, was proposed for semantic segmentation, with two symmetric paths named the contracting path and the expanding path (Ronneberger et al., 2015). The contracting path extracts hierarchical features by gradually downsampling the spatial resolution of the feature maps, while the expanding path learns more contextual information by progressively restoring the spatial resolution. Subsequently, the encoder-decoder framework has become the standard structure of remote sensing image segmentation networks (Badrinarayanan et al., 2017; Chen et al., 2018a) (Sun et al., 2019). Based on encoder-decoder structure, (Diakogiannis et al., 2020; Yue et al., 2019; Zhou et al., 2018) designed different skip connections to capture more abundant context, while (Liu et al., 2018; Zhao et al., 2017b) (Shen et al., 2019) developed various decoders to retain semantic information.

The encoder-decoder CNN-based methods, although have achieved encouraging performance, encounter bottlenecks in urban scene interpretation (Sherrah, 2016) (Marmanis et al., 2018; Nogueira et al., 2019). To be specific, CNN-based segmentation networks with limited receptive fields can only extract local semantic features and lack the capability to model the global information from the whole image. However, within fine-resolution remotely sensed urban scene images, complicated patterns and human-made objects occur frequently (Kampffmeyer et al., 2016; Marcos et al., 2018) (Audebert et al., 2018). It is difficult to identify these complex objects if only relying on the local infromation.

## 2.2 Global contextual information modelling

To liberate the network from the local pattern focus of CNNs, many attempts have been conducted to modelling global contextual information, while the most popular way is incorporating attention mechanisms into networks. For example, Wang et al. modified the dot-product self-attention mechanism and applied it to computer vision domains (Wang et al., 2018). Fu et al. appended two types of attention modules on top of a dilated FCN to adaptively integrate local features with their global dependencies (Fu et al., 2019). Huang et al. proposed a criss-cross attention block to aggregate informative global features (Huang et al., 2020). Yuan et al. developed an object context block to explore object-based global relations (Yuan et al., 2020).

Attention mechanisms also improve the performance of remote sensing image segmentation networks. Yang et al. proposed an attention-fused network to fuse high-level and low-level semantic features and obtain state-of-the-art results in the semantic segmentation of fine-resolution remote sensing images (Yang et al., 2021b). Li et al. integrated lightweight spatial and channel attention modules to refine semantic features adaptively for high-resolution remotely sensed image segmentation (Li et al., 2020). Ding et al. designed a local attention block with an embedding module to capture richer contextual information (Ding et al., 2021). Li et al. developed a linear attention mechanism to reduce the computational complexity while improving performance (Li et al., 2021a). However, the above attention modules restrict the global feature representation due to over-reliance on convolutional operations. Furthermore, a single attention module cannot model the global information at multi-level semantic features in the decoder.

## 2.3 Transformer-based semantic segmentation methods

Recently, several attempts were made to apply the Transformer for global information extraction (Vaswani et al., 2017). Different from the CNN structure, the Transformer translates 2D image-based tasks into 1D sequence-based tasks. Due to the powerful sequence-to-sequence modelling ability, the Transformer demonstrates superior characterization of extracting global context than the above-mentioned attention-alone models and obtains state-of-the-art results on fundamental vision tasks, such as image classification (Dosovitskiy et al., 2020), object detection (Zhu et al., 2020) and semantic segmentation (Zheng et al., 2021). Driven by this, many researchers in the remote sensing field have applied the Transformer for remote sensing image scene classification (Bazi et al., 2021; Deng et al., 2021), hyperspectral image classification (Hong et al., 2021) (He et al., 2021), object detection (Li et al., 2022a), change detection (Chen et al., 2021a), and especially semantic segmentation (Wang et al., 2022; Wang et al., 2021b).

Most of the existing Transformers for semantic segmentation still follow the encoder-decoder framework. According to different encoder-decoder combinations, they can be divided into two categories. The first is constructed by a Transformer-based encoder and a Transformer-based decoder, namely the pure Transformer structure. Typical models include the Segmenter (Strudel et al., 2021), SegFormer (Xie et al., 2021) and SwinUNet (Cao et al., 2021). The second adopts a hybrid structure, which is composed of a Transformer-based encoder and a CNN-based decoder. Transformer-based semantic segmentation methods commonly follow the second structure. For example, the TransUNet employed the hybrid vision Transformer (Dosovitskiy et al., 2020) as the encoder for stronger feature extraction and obtains state-of-the-art results in medical image segmentation (Chen et al., 2021b). The DC-Swin introduced Swin Transformer (Liu et al., 2021) as the encoder and designs a densely connected convolutional decoder for fine-resolution remote sensing image segmentation, surpassing the CNN-based methods by a large gap (Wang et al., 2022). (Panboonyuen et al., 2021) also selected the Swin Transformer as the encoder and utilizes various CNN-based decoders, such as UNet (Ronneberger et al., 2015), FPN (Kirillov et al., 2019) and PSP (Zhao et al., 2017a), for semantic segmentation of remotely sensed images, obtaining advanced accuracy.

Despite the above advantages, the computational complexity of the Transformer-based encoder is much higher than the CNN-based encoder due to its square-complexity self-attention mechanism (Vaswani et al., 2017), which seriously affects its potential and feasibility for urban-related real-time applications. Thus, to fully harness the global context extraction ability of Transformers without resulting in high computational complexity, in this paper, we present a UNet-like Transformer with a CNN-based encoder and a Transformer-based decoder for efficient semantic segmentation of remotely sensed urban scene images. Specifically, for our UNetFormer, we select the lightweight backbone, i.e. ResNet18, as the encoder and develop an efficient global-local attention mechanism to construct Transformer blocks in the decoder. The proposed efficient global-local attention mechanism adopts a dual-branch structure, i.e. a global branch and a local branch. Such a structure allows the attention block to capture both global and local contexts, thereby surpassing the single-branch efficient attention mechanisms in Transformers that only capture global contexts (Liu et al., 2021; Zhang and Yang, 2021).

## 3. Methodology

As illustrated in **Fig. 3,** the proposed UNetFormer is constructed using a CNN-based encoder and a Transformer-based decoder. A detailed description of each component is given in the following sections.

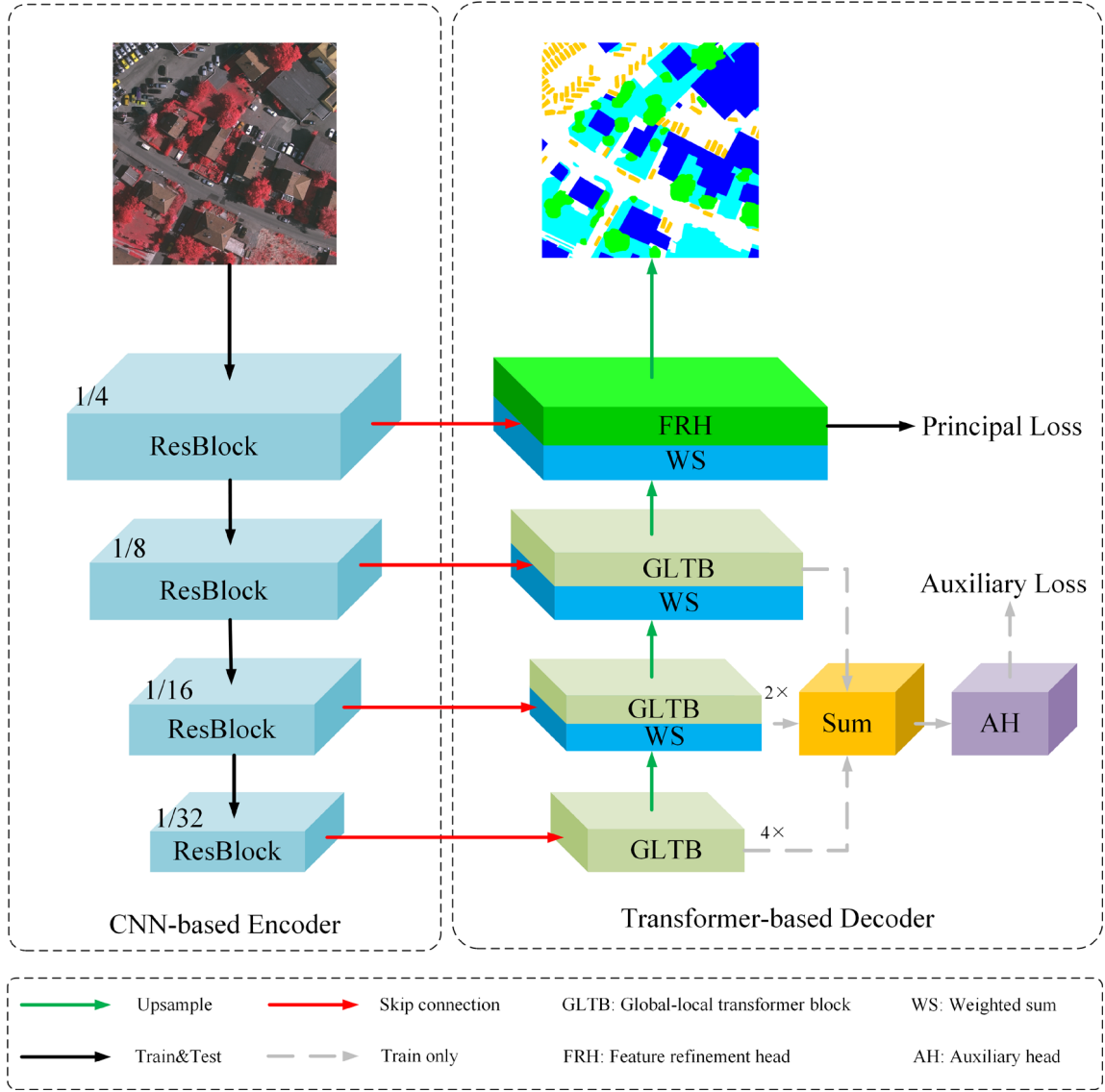

**Fig. 3.** An overview of the UNetFormer.

## 3.1 CNN-based encoder

As the ResNet18 (He et al., 2016) has demonstrated effectiveness and efficiency simultaneously in a wide range of real-time semantic segmentation tasks, we select the pre-trained ResNet18 as the encoder here to extract multi-scale semantic features with significantly low computational cost. ResNet18 consists of four-stage Resblocks, with each stage down-sampling the feature map with a scale factor of 2. In the proposed UNetFormer, the feature maps generated by each stage are fused with the corresponding feature maps of the decoder by a $1 \times 1$ convolution with the channel dimension in 64, i.e., the skip connection. Specifically, the semantic features produced by the Resblocks are aggregated with the features generated by the GLTB of the decoder using a weighted sum operation. The weighted sum operation weights the two features selectively based on their contributions to segmentation accuracy, thereby learning more generalized fusion features (Tan et al., 2020). The formulation of the weighted sum operation can be denoted as:

$$\mathbf{FF} = \alpha \cdot \mathbf{RF} + (1 - \alpha) \cdot \mathbf{GLF} \tag{1}$$

where $\mathbf{FF}$ represents the fused feature, $\mathbf{RF}$ denotes the feature produced by the Resblocks, and $\mathbf{GLF}$ indicates the feature generated by the global-local Transformer block.

## 3.2 Transformer-based decoder

Complicated human-made objects occur frequently in fine-resolution remotely sensed urban images, which makes it difficult to achieve precise real-time segmentation without global semantic information. To capture the global context, mainstream solutions focus on attaching a single attention block at the end of the network (Wang et al., 2018) or introducing Transformers as the encoder (Chen et al., 2021b). The former cannot capture multi-scale global features, whereas the latter significantly increases the complexity of the network and loses spatial details. In contrast, in the proposed UNetFormer, we utilize three global-local Transformer blocks and a feature refinement head to build a lightweight Transformer-based decoder, as shown in **Fig. 3**. With such a hierarchical and lightweight design, the decoder is capable of capturing both global and local contexts at multiple scales while maintaining high efficiency.

### 3.2.1 Global-local Transformer block (GLTB)

The global-local Transformer block consists of the global-local attention, multilayer perceptron, two batch normalization layers and two additional operations, as shown in **Fig. 1 (b)**.

**Efficient Global-local attention**: Although the global context is crucial for semantic segmentation of complex urban scenes, local information is still essential to preserve rich spatial details. In this regard, the proposed efficient global-local attention constructs two parallel branches to extract the global and local contexts, respectively, as shown in **Fig. 4 (a)**.

As a relatively shallow structure, the local branch employs two parallel convolutional layers with kernel sizes of 3 and 1 to extract the local context. Two batch normalization operations are then attached before the final sum operation.

The global branch deploys the window-based multi-head self-attention to capture global context. As illustrated in **Fig 4. (b)**, we first use a standard $1 \times 1$ convolution to expand the channel dimension of the input 2D feature map $\in \mathbb{R}^{B \times C \times H \times W}$ to three times. Then, we apply the window partition operation to split the 1D sequence $\in \mathbb{R}^{\left(3 \times B \times \frac{H}{w} \times \frac{W}{w} \times h\right) \times (w \times w) \times \frac{C}{h}}$ into the query (Q), key (K) and value (V) vectors. The channel dimension $C$ is set to 64. The window size $w$ and the number of heads $h$ are both set to 8. The details of the window-based multi-head self-attention can refer to Swin Transformer (Liu et al., 2021).

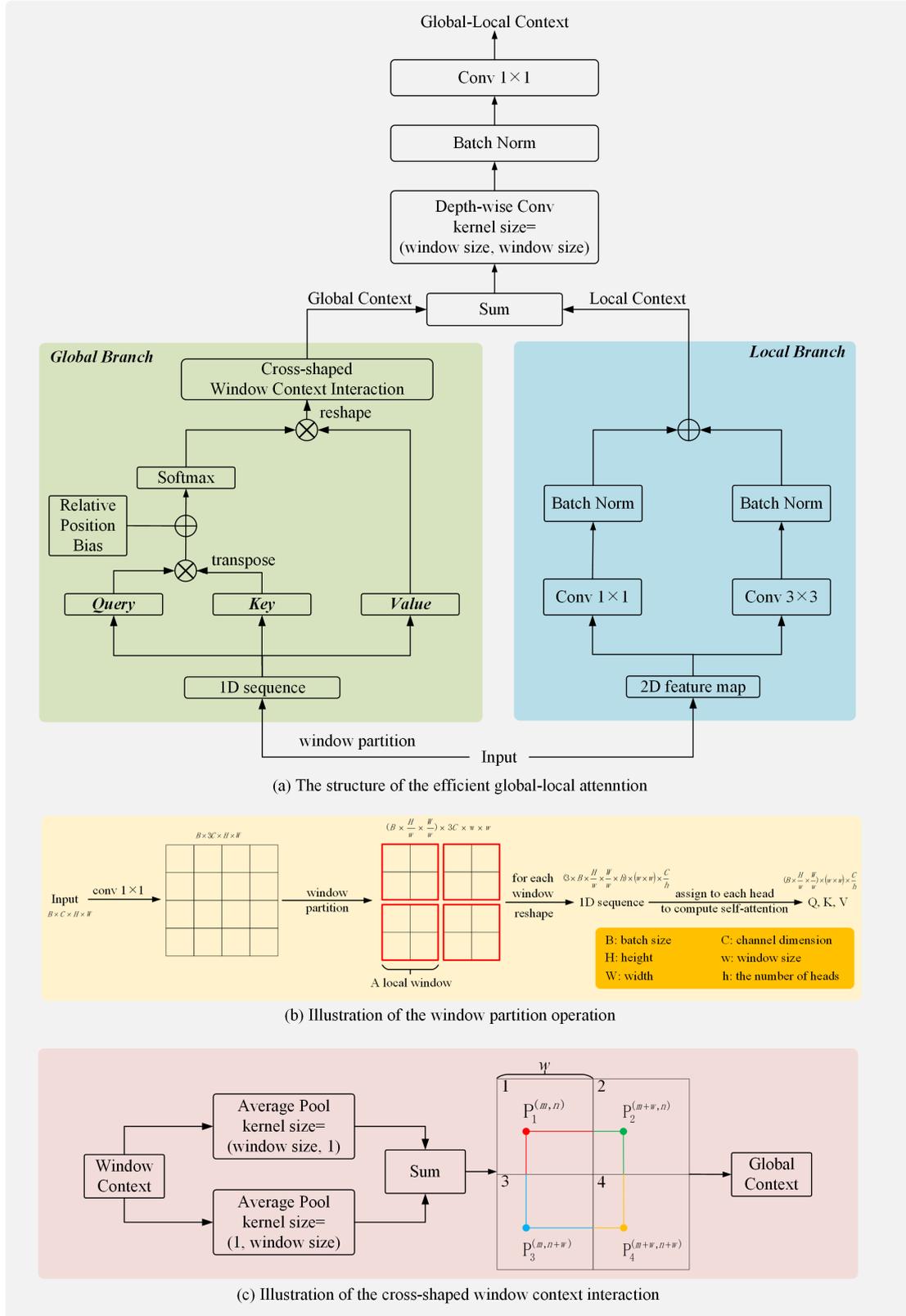

**Fig. 4.** Illustration of the efficient global-local attention.

Performing self-attention in a non-overlapping local window, although being efficient, can destroy the spatial consistency of urban scenes due to the lack of interactions across windows. The Swin Transformer introduces an extra shifted Transformer block to mine the relationship between local windows. Although the ability to capture cross-window relations increases, the computation significantly surges accordingly. In this paper, we propose a cross-shaped window context interaction module to capture the cross-window relations with high computational efficiency. As illustrated in **Fig. 4 (c)**, the cross-shaped window context interaction module fuses the two feature maps produced by a horizontal average pooling layer and a vertical average pooling layer, thereby capturing the global context. Specifically, the horizontal average pool layer establishes the horizontal relationship between Windows, such as $Win_1 = H(Win_2)$. For any point $P_1^{(m,n)}$ in Window 1, its dependency with $P_2^{(m+w,n)}$ in Window 2 can be modelled as:

$$P_1^{(m,n)} = \frac{\sum_{i=0}^{w-m-1} P_1^{(m+i,n)} + \sum_{j=0}^{m} P_2^{(m+w-j,n)}}{w} \tag{2}$$

$$P_1^{(m+i,n)} = D_i\left(P_1^{(m,n)}\right) \tag{3}$$

$$P_2^{(m+w-j,n)} = D_j\left(P_2^{(m+w,n)}\right) \tag{4}$$

$$P_1^{(m,n)} = \frac{\sum_{i=0}^{w-m-1} D_i\left(P_1^{(m,n)}\right) + \sum_{j=0}^{m} D_j\left(P_2^{(m+w,n)}\right)}{w} \tag{5}$$

Where $w$ is the window size. $D$ denotes the self-attention computation, which can model dependencies of pixel pairs in a local window. Thus, for any other point $P_1^{(m+i,n)}$ in the red path of Window 1, its dependency with $P_1^{(m,n)}$ can be modelled by Eq.(3). For any other point $P_2^{(m+w-j,n)}$ in the green path of Window 2, its dependency with $P_2^{(m+w,n)}$ can be modelled by Eq.(4). Eq.(2) can be rewritten as Eq.(5), i.e. the dependency between $P_1^{(m,n)}$ and $P_2^{(m+w,n)}$ is modelled. Based on this cross-window pixel-wise dependency, the horizontal relationship between Windows 1 and 2 can be established. Similarly, the vertical relationship between Window 1 and 3 can be established in the same way, i.e. $Win_1 = V(Win_3)$, and for Window 4, $Win_1 = V(H(Win_4)) + H(V(Win_4))$. Generalized to an M×M input (M denotes the number of windows), by connecting more intermedia windows like Window 2 and Window 3, the long-range dependency between any two windows can be modelled. Thus, the cross-shaped window context interaction module can model the window-wise long-range dependencies, thereby capturing the global context.

Besides, the global context in the global branch is further aggregated with the local context in the local branch to produce the global-local context. Finally, we employ a depth-wise convolution, a batch normalization operation and a standard 1×1 convolution to characterize the fine-grained global-local context.

### 3.2.2 Feature refinement head (FRH)

The shallow feature produced by the first Resblock preserves rich spatial details of urban scenes, but lacks semantic content, while the deep global-local feature provides precise semantic information, but with a coarse spatial resolution. Hence, a direct sum operation on these two features, although fast, can reduce segmentation accuracy (Poudel et al., 2018; Poudel et al., 2019; Yu et al., 2018). In this paper, we develop a feature refinement head to shrink the semantic gap between the two features for further accuracy improvement.

As can be seen in **Fig. 5**, we perform a weighted sum operation on the two features first to take full advantage of the precise semantic information and spatial details. The fused feature is then selected as the input of the FRH, as shown in **Fig. 3**. Second, we construct two paths to strengthen the channel-wise and spatial-wise feature representation. Specifically, the channel path employs a global average pooling layer to generate a channel-wise attentional map $C \in \mathbb{R}^{1 \times 1 \times c}$, where $c$ denotes the channel dimension. The reduce & expand operation contains two $1 \times 1$ convolutional layers, which first reduces the channel dimension $c$ by a factor of 4 and then expands it to the original. The spatial path utilizes a depth-wise convolution to produce a spatial-wise attentional map $S \in \mathbb{R}^{h \times w \times 1}$, where $h$ and $w$ represent the spatial resolution of the feature map. The attentional features generated by the two paths are further fused using a sum operation. Finally, a post-processing $1 \times 1$ convolutional layer and an upsampling operation are applied to produce the final segmentation map. Notably, a residual connection is introduced to prevent network degradation.

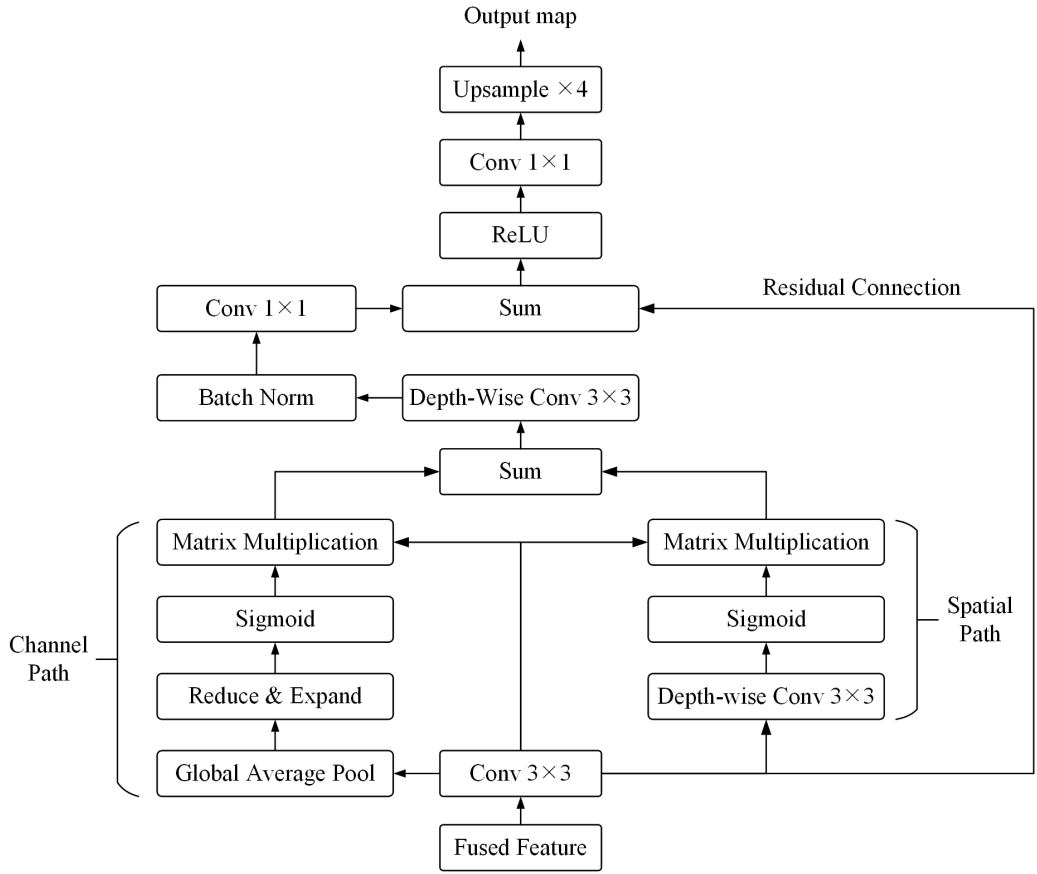

**Fig. 5.** The feature refinement head.

### 3.3 Loss function

In the training phase, we employ not only the primary feature refinement head but also build an extra auxiliary head to optimize the global-local Transformer blocks, as shown in **Fig. 3**. This multi-head segmentation architecture has been demonstrated to be effective in previous research (Yu et al., 2020; Zhu et al., 2019). Based on the multi-head design, we apply a principal loss and an auxiliary loss to train the entire network. The principal loss $\mathcal{L}_p$ is a combination of a dice loss $\mathcal{L}_{dice}$ and a cross-entropy loss $\mathcal{L}_{ce}$, which can be formulated as:

$$\mathcal{L}_{ce} = -\frac{1}{N}\sum\nolimits_{n=1}^{N}\sum\nolimits_{k=1}^{K} y_k^{(n)}\log\hat{y}_k^{(n)} \tag{6}$$

$$\mathcal{L}_{dice} = 1 - \frac{2}{N}\sum\nolimits_{n=1}^{N}\sum\nolimits_{k=1}^{K} \frac{\hat{y}_k^{(n)}y_k^{(n)}}{\hat{y}_k^{(n)} + y_k^{(n)}} \tag{7}$$

$$\mathcal{L}_p = \mathcal{L}_{ce} + \mathcal{L}_{dice} \tag{8}$$

where $N$ and $K$ denote the number of samples and the number of categories, respectively. $y^{(n)}$ and $\hat{y}^{(n)}$ represent the one-hot encoding of the true semantic labels and the corresponding softmax output of the network, $n \in [1, \cdots, N]$. $\hat{y}_k^{(n)}$ is the confidence of sample $n$ belonging to the category $k$. We select the cross-entropy loss as the auxiliary loss $\mathcal{L}_{aux}$ and deploy it on the auxiliary head. The auxiliary head takes the fused feature of the three global-local Transformer blocks as the input and constructs a $3\times3$ convolution layer with batch normalization and ReLU, a $1\times1$ convolution layer and an upsampling operation to generate the output. For a better combination with the principle loss, the auxiliary is further multiplied by a factor $\alpha$. Thus, the overall loss $\mathcal{L}$ can be formulated as:

$$\mathcal{L} = \mathcal{L}_p + \alpha \times \mathcal{L}_{aux} \qquad (9)$$

where $\alpha$ is set to 0.4 by default.

## 4. EXPERIMENTS

### 4.1 Experimental settings

#### 4.1.1 Datasets

**UAVid**: As a fine-resolution Unmanned Aerial Vehicle (UAV) semantic segmentation dataset, the UAVid dataset focuses on urban street scenes with two spatial resolutions ($3840 \times 2160$ and $4096 \times 2160$) and eight classes (Lyu et al., 2020). Segmentation of UAVid is challenging due to the fine spatial resolution of images, heterogeneous spatial variation, vague categories and generally complex scenes. To be specific, there are 42 sequences with a total of 420 images in the dataset, where 200 images are used for training, 70 images for validation and the officially provided 150 images for testing. In our experiments, each image was padded and cropped into eight $1024 \times 1024$ px patches.

**Vaihingen**: The Vaihingen dataset consists of 33 very fine spatial resolution TOP image tiles at an average size of $2494 \times 2064$ pixels. Each TOP image tile has three multispectral bands (near infrared, red, green) as well as a digital surface model (DSM) and normalized digital surface model (NDSM) with a 9 cm ground sampling distance (GSD). The dataset involves five foreground classes (impervious surface, building, low vegetation, tree, car) and one background class (clutter). In our experiments, only the TOP image tiles were used without the DSM and NDSM. And we utilized ID: 2, 4, 6, 8, 10, 12, 14, 16, 20, 22, 24, 27, 29, 31, 33, 35, 38 for testing, and the remaining 16 images for training. The image tiles were cropped into 1024×1024 px patches.

**Potsdam**: The Potsdam dataset contains 38 very fine spatial resolution TOP image tiles (GSD 5 cm) at a size of 6000×6000 pixels and involves the same category information as the Vaihingen dataset. Four multispectral bands (red, green, blue, and near infrared), as well as the DSM and NDSM, are provided in the dataset. We utilized ID: 2_13, 2_14, 3_13, 3_14, 4_13, 4_14, 4_15, 5_13, 5_14, 5_15, 6_13, 6_14, 6_15, 7_13 for testing, and the remaining 23 images (except image 7_10 with error annotations) for training. Similarly, only three bands (red, green, blue) were utilized and the original image tiles were cropped into 1024×1024 px patches in the experiments.

**LoveDA**: The LoveDA dataset contains 5987 fine-resolution optical remote sensing images (GSD 0.3 m) at a size of 1024×1024 pixels and includes 7 landcover categories, i.e. building, road, water, barren, forest, agriculture and background (Wang et al., 2021a). Specifically, 2522 images are used for training, 1669 images for validation and the officially provided 1796 images for testing. The dataset encompasses two scenes (urban and rural) which are collected from three cities (Nanjing, Changzhou and Wuhan) in China. Therefore, considerable challenges are brought due to the multi-scale objects, complex background and inconsistent class distributions.

### 4.1.2 Implementation Details

All models in the experiments were implemented with the PyTorch framework on a single NVIDIA GTX 3090 GPU. For fast convergence, we deployed the AdamW optimizer to train all models in the experiments. The base learning rate was set to 6e-4. The cosine strategy was employed to adjust the learning rate.

For the UAVid dataset, random vertical flip, random horizontal flip and random brightness were used to the input in the size of 1024×1024 for data augmentation in the training period, while the training epoch was set as 40 and the batch size was 8. In the test procedure, the test-time augmentation (TTA) strategies like vertical flip and horizontal flip were used.

For the Vaihinge, Potsdam and LoveDA datasets, the images were randomly cropped into 512×512 patches. For training, the augmentation techniques like random scale ([0.5, 0.75, 1.0, 1.25, 1.5]), random vertical flip, random horizontal flip and random rotate were adopted during the training process, while the training epoch was set as 100 and the batch size was 16. During the test phase, multi-scale and random flip augmentations were used.

### 4.1.3 Evaluation metrics

The evaluation metrics used in our experiments included two major categories. The first one was to evaluate the accuracy of the network including the overall accuracy (OA), mean F1 score (F1) and mean intersection over union (mIoU). The second one was to evaluate the scale of the network, including the floating point operation count (Flops) to evaluate the complexity, the frames per second (FPS) to evaluate the speed, the memory footprint (MB) and the number of model parameters (M) to evaluate the memory requirement.

### 4.1.4 Models for comparison

We selected a comprehensive set of benchmark methods for quantitative comparison including

(i)     CNN-based lightweight networks developed for efficient semantic segmentation: context aggregation network (CANet) (Yang et al., 2021a), bilateral segmentation network (BiSeNet) (Yu et al., 2018), ShelfNet (Zhuang et al., 2019), SwiftNet (Oršić and Šegvić, 2021), Fast-SCNN (Poudel et al., 2019), DABNet (Li et al., 2019), ERFNet (Romera et al., 2017) and ABCNet (Li et al., 2021c).

(ii)     CNN-based attentional networks: dual attention network (DANet) (Fu et al., 2019), fast attention network (FANet) (Hu et al., 2020), local attention network (LANet) (Ding et al., 2021), criss-cross network (CCNet) (Huang et al., 2020), multi-stage attention residual UNet (MAResU-Net) (Li et al., 2021a) and multi-attention network (MANet) (Li et al., 2021b),

(iii)    CNN-based networks for semantic segmentation of remote sensing images: DST_5 (Sherrah, 2016), V-FuseNet (Audebert et al., 2018), CASIA2 (Liu et al., 2018), DLR_9 (Marmanis et al., 2018), RoteEqNet (Marcos et al., 2018), UFMG_4 (Nogueira et al., 2019), HUSTW5 (Sun et al., 2019), TreeUNet (Yue et al., 2019), ResUNet-a (Diakogiannis et al., 2020), S-RA-FCN (Mou et al., 2020), DDCM-Net (Liu et al., 2020), EaNet (Zheng et al., 2020a), HMANet (Niu et al., 2021) and AFNet (Yang et al., 2021b),

(iv)     hybrid Transformer-based networks with a Transformer-based encoder and a CNN-based decoder: TransUNet (Chen et al., 2021b), SwinUperNet (Liu et al., 2021), DC-Swin (Wang et al., 2022), STranFuse (Gao et al., 2021), SwinB-CNN+BD (Zhang et al., 2022), SwinTF-FPN (Panboonyuen et al., 2021), BANet (Wang et al., 2021b), CoaT (Xu et al., 2021), BoTNet (Srinivas et al., 2021) and ResT (Zhang and Yang, 2021),

(v)     fully Transformer-based networks with a Transformer-based encoder and a Transformer-based decoder: SwinUNet (Cao et al., 2021), SegFormer (Xie et al., 2021) and Segmenter (Strudel et al., 2021).

## 4.2 Ablation study

### 4.2.1 Each component of UNetFormer

To evaluate the performance of each component of the proposed UNetFormer separately, we conducted a series of ablation experiments on the UAVid, Vaihingen and Potsdam datasets. For a fair comparison, the test time augmentation strategies and auxiliary loss were not used in all ablation studies. The results are illustrated in **TABLE 1**.

**TABLE 1**. Ablation study of each component of the UNetFormer.

| Dataset | Method | mIoU |
|---------|--------|------|
| UAVid | Baseline | 65.4 |
| | Baseline+GLTB-SUM | 67.8 |
| | Baseline+GLTB | 68.8 |
| | Baseline+GLTB+FRH | 70.0 |
| Vaihingen | Baseline | 77.1 |
| | Baseline+GLTB-SUM | 79.4 |
| | Baseline+GLTB | 80.6 |
| | Baseline+GLTB+FRH | 81.6 |
| Potsdam | Baseline | 82.5 |
| | Baseline+GLTB-SUM | 83.8 |
| | Baseline+GLTB | 84.9 |
| | Baseline+GLTB+FRH | 85.5 |

**Baseline**: The baseline was constructed by the U-Net with a ResNet18 backbone, which only models the local contextual information in the decoder.

**The global-local Transformer block (GLTB)**: Three global-local Transformer blocks were incorporated into the baseline to build the **Baseline+GLTB**. Meanwhile, to illustrate the contribution of the cross-shaped window context interaction module in the GLTB, we remove it and apply a direct sum operation on the window context and local context, thereby constructing a simple variant **Baseline+GLTB-SUM**. As shown in **TABLE 1**, the deployment of GLTB provides a significant increase of mIoU by 3.4% on the UAVid validation set, where the contribution of the cross-shaped window context interaction module to increase accuracy is 1.0%. Meanwhile, **Baseline+GLTB** achieves an increase of greater than 2.4% in mIoU on the Vaihingen and Potsdam test sets, where the increase provided by the cross-shaped window context interaction module is 1.2% and 1.1%, respectively. To sum up, the results not only demonstrate the effectiveness of GLTB but also indicate the necessity of applying the cross-shaped window context interaction module.

**The feature refinement module (FRH)**: We inserted the feature refinement head into **Baseline+GLTB** to generate the entire UNetFormer (indicated as **Baseline+GLTB+FRH**). As shown in **TABLE 1**, with the employment of FRH, the mIoU is boosted by 1.0% at least, demonstrating the validity of the proposed feature refinement module.

### 4.2.2 Efficient global-local attention

To demonstrate the advantages of the proposed efficient global-local attention, we replaced it with other advanced attention mechanisms to reconstruct the variants of UNetformer for ablation studies. Benefiting from the dual-branch structure and the captured global-local context, the deployment of our global-local attention achieves the highest mIoU (70.0%) on the UAVid validation set, as listed in **TABLE 2**. Besides, the proposed global-local attention also demonstrates superiority in terms of complexity, memory requirement, parameters and inference speed. Especially, our method is more accurate and faster than the efficient attention mechanisms in Transformers, i.e. the shifted window attention and the efficient multi-head self-attention.

**TABLE 2**. Ablation studies of different attention mechanisms on the UAVid dataset. We report the speed with an input size of 1024×1024 on a single NVIDIA GTX 3090 GPU. The best values in the column are in bold.

| Attention mechanism | Complexity(G) | Memory(MB) | Parameters(M) | Speed(FPS) | mIoU |
|---|---|---|---|---|---|
| Dual attention (Fu et al., 2019) | 68.9 | 2416.4 | 12.6 | 53.8 | 67.3 |
| Criss-cross attention (Huang et al., 2020) | 67.2 | 1318.4 | 12.4 | 79.9 | 68.3 |
| Linear attention (Li et al., 2021b) | 67.8 | 1339.5 | 12.5 | 91.5 | 69.0 |
| Patch attention (Ding et al., 2021) | 66.8 | 1320.5 | 12.3 | 95.7 | 68.9 |
| Efficient multi-head self-attention (Zhang and Yang, 2021) | 67.5 | 2444.2 | 12.5 | 63.6 | 67.9 |
| Shifted window attention (Liu et al., 2021) | 72.7 | 1652.0 | 13.1 | 67.0 | 68.5 |
| Efficient global-local attention (Sudre et al.) | **46.9** | **1003.8** | **11.7** | **115.6** | **70.0** |

### 4.2.3 Network stability

To evaluate the network stability, we trained the UNetFormer with different input sizes, including square inputs like 512×512, 1024×1024 and 2048×2048 as well rectangular inputs like 512×1024 and 1024×2048. From the experimental results in **TABLE 3**, the UNetFormer demonstrates stability when performing different input sizes, while the deviation of the mIoU is less than 0.7%. The middle input size of 1024×1024 obtains the best mIoU on the UAVid validation set. Furthermore, the square inputs yield relatively higher scores than the rectangular inputs, and too large input size like 2048×2048 can reduce the IoU of very small object "human".

**TABLE 3** Ablation studies of different input sizes on the UAVid dataset.

| Input size | Clutter | Building | Road | Tree | Vegetation | MovingCar | StaticCar | Human | mIoU |
|---|---|---|---|---|---|---|---|---|---|
| 512×512 | 63.1 | 90.7 | 76.4 | 77.4 | 68.1 | 70.3 | 65.9 | 46.2 | 69.8 |
| 512×1024 | 61.9 | 91.0 | 74.9 | 76.9 | 69.1 | 70.4 | 65.6 | 44.3 | 69.3 |
| 1024×1024 | 63.6 | 91.2 | 76.4 | 77.7 | 68.2 | 71.6 | 66.1 | 44.8 | 70.0 |
| 1024×2048 | 63.0 | 91.2 | 76.2 | 77.5 | 68.7 | 69.8 | 65.2 | 44.6 | 69.5 |
| 2048×2048 | 63.4 | 91.2 | 76.0 | 77.9 | 70.1 | 70.4 | 65.7 | 42.5 | 69.7 |

### 4.2.4 Encoder choice

Current Transformer-based segmentation networks commonly apply the Transformer as the encoder. This choice, although has been justified for accurate semantic information, reduces the execution speed of the network significantly, which is not suitable for real-time applications. To demonstrate it, we replace our ResNet18 encoder with lightweight Transformers, i.e. ViT-Tiny (Dosovitskiy et al., 2020), Swin-Tiny (Liu et al., 2021) and CoaT-Mini (Xu et al., 2021), for ablation studies (**TABLE 4**). The results reveal that introducing lightweight Transformers as the encoder provides a limited improvement of accuracy (within 0.6% in mIoU) but reduces the inference speed of the UNetFormer seriously. Thus, for real-time urban scene segmentation, the application of a lightweight CNN-based encoder like ResNet18 is the currently best scheme.

**TABLE 4** Ablation studies of different encoders on the UAVid dataset. The complexity and speed are measured by a 1024×1024 input on a single NVIDIA GTX 3090 GPU.

| Method | Encoder | Complexity(G) | Parameters(M) | Speed(FPS) | mIoU |
|---|---|---|---|---|---|
| UNetFormer | ViT-Tiny (Dosovitskiy et al., 2020) | 35.31 | 8.6 | 30.2 | 69.1 |
| | Swin-Tiny (Liu et al., 2021) | 104.4 | 28.0 | 28.8 | 70.6 |
| | CoaT-Mini (Xu et al., 2021) | 159.7 | 10.6 | 10.6 | 70.5 |
| | ResNet18 | 46.9 | 11.7 | 115.6 | 70.0 |

### 4.2.5 Encoder-decoder combination

To illustrate the superiority of our hybrid structure for efficient semantic segmentation, we selected the UNet, SwinUNet and TransUNet for ablation experiments on the UAVid dataset. Since the SwinUNet requires huge GPU memory, the input size was all set as 512×512 for training. The results from **TABLE 5** reveal that the proposed UNetFormer exceeds the compared networks significantly in terms of complexity and speed while providing a competitive accuracy on the UAVid validation set. Specifically, in comparison with the UNet constructed by the pure CNN structure, the UNetFormer achieves an increase of 4.3% in mIoU. Compared to the pure Transformer network SwinUNet, the UNetFormer saves 80% of the computational complexity. Although the TransUNet constructed by a Transformer-based encoder and a CNN-based decoder surpasses ours by 0.5% in mIoU, it is 7 times slower and has much more parameters due to its heavy and complicated Transformer-based encoder. For real-time urban application scenarios, the high execution speed and lightweight model volume are much more important than the slight accuracy reduction. Thus, in comparison with other combinations, the advantage of our hybrid structure, i.e. CNN-based encoder and Transformer-based decoder, is significant.

**TABLE 5** Ablation studies of different encoder-decoder combinations on the UAVid dataset. The complexity and speed are measured by a 512×512 input on a single NVIDIA GTX 3090 GPU.

| Method | Backbone | Encoder | Decoder | Complexity(G) | Memory(MB) | Parameters(M) | Speed(FPS) | mIoU |
|--------|----------|---------|---------|---------------|------------|---------------|-----------|------|
| UNet (Ronneberger et al., 2015) | - | CNN | CNN | 184.6 | 1622.0 | 31.0 | 50.9 | 65.5 |
| SwinUNet (Cao et al., 2021) | Swin-Tiny | Transformer | Transformer | 237.4 | 2001.5 | 41.4 | 46.9 | 68.3 |
| TransUNet (Chen et al., 2021b) | ViT-R50 | Transformer | CNN | 233.7 | 1245.7 | 90.7 | 43.2 | **70.3** |
| UNetFormer | ResNet18 | CNN | Transformer | **11.7** | **250.9** | **11.7** | **322.4** | 69.8 |

## 4.3 Experiment results

### 4.3.1 Comparison of network efficiency

Complexity and speed are critical for evaluating a network, especially in real-time urban applications. We compared our UNetFormer with efficient segmentation networks based on the mIoU, GPU memory footprint, complexity, parameters and speed on the official UAVid test set. The comparison results are listed in **Table 6**. In comparison with the fastest and most shallow model Fast-SCNN, the proposed UNetFormer outperforms it by a large margin of 21.0% in mIoU. In comparison with the state-of-the-art CNN-based models of the same volume, our UNetFormer achieves a competitive inference speed of 115.6 FPS, while surpassing other networks by more than 4.0% in mIoU. Notably, our method exceeds the advanced hybrid Transformer network CoaT by 2.0% in mIoU while being 10 times faster. Meanwhile, the proposed method outperforms the pure Transformer network Segmenter by 9.1% in mIoU while being 7 times faster. The outstanding trade-off between accuracy and speed demonstrates the efficiency of our hybrid structure and the effectiveness of the proposed GLTB and FRH.

**TABLE 6**. Quantitative comparison results on the UAVid test set with state-of-the-art lightweight networks. The complexity and speed are measured by a 1024×1024 input on a single NVIDIA GTX 3090 GPU.

| Method | Backbone | Memory(MB) | Parameters(M) | Complexity(G) | Speed | mIoU |
|---|---|---|---|---|---|---|
| Fast-SCNN (Poudel et al., 2019) | - | 619.4 | 1.1 | 3.4 | 222.7 | 45.9 |
| Segmenter (Strudel et al., 2021) | ViT-Tiny | 828.6 | 6.7 | 26.8 | 14.7 | 58.7 |
| BiSeNet (Yu et al., 2018) | ResNet18 | 970.6 | 12.9 | 51.8 | 121.9 | 61.5 |
| DANet (Fu et al., 2019) | ResNet18 | 611.1 | 12.6 | 39.6 | 189.4 | 60.6 |
| FANet (Hu et al., 2020) | ResNet18 | 971.9 | 13.6 | 86.8 | 94.9 | - |

| | | | | | |
|---|---|---|---|---|---|
| ShelfNet (Zhuang et al., 2019) | ResNet18 | 579.0 | 14.6 | 46.7 | 141.4 | 47.0 |
| SwiftNet (Oršić and Šegvić, 2021) | ResNet18 | 835.8 | 11.8 | 51.6 | 138.7 | 61.1 |
| MANet (Li et al., 2021b) | ResNet18 | 1169.2 | 12.0 | 51.7 | 75.6 | 62.6 |
| ABCNet (Li et al., 2021c) | ResNet18 | 1105.1 | 14.0 | 62.9 | 102.2 | 63.8 |
| SegFormer (Xie et al., 2021) | MiT-B1 | 933.2 | 13.7 | 63.3 | 31.3 | 66.0 |
| BoTNet (Srinivas et al., 2021) | ResNet18 | 710.5 | 17.6 | 49.9 | 135.0 | 63.2 |
| CoaT (Xu et al., 2021) | CoaT-Mini | 3133.8 | 11.1 | 104.8 | 10.6 | 65.8 |
| UNetFormer | ResNet18 | 1003.7 | 11.7 | 46.9 | 115.6 | 67.8 |

## 4.3.2 Results on the UAVid dataset

UAVid is a large-scale urban scene segmentation dataset, where the images are captured by unmanned aerial vehicles in different cities and under different lighting conditions. Thus, it is challenging to obtain high scores on this dataset. We trained several advanced efficient segmentation networks and provide a detailed comparison of results on the official UAVid test set. As illustrated in **Table 7**, our method yields the best mIoU (67.8%) while maintaining the advantages in the per-class IoU. Specifically, the proposed UNetFormer not only exceeds the efficient CNN-based network ABCNet by 4.0% in mIoU but also outperforms the recent hybrid Transformer-based networks BANet and BoTNet by 3.2% and 4.6%, respectively. Particularly, the "human" class is hard to handle since it is an extremely small object. Nonetheless, the IoU of this class achieved by our UNetFormer is at least 8.6% higher than for other methods. Furthermore, the segmentation results from the UAVid validation set (**Fig. 6**) and the visualization results from the UAVid test set (**Fig. 7**) also demonstrate the effectiveness of our UNetFormer.

**TABLE 7**. Quantitative comparison of results on the UAVid test set with state-of-the-art lightweight models. The best values in the column are in bold.

| Method | Backbone | Clutter | Building | Road | Tree | Vegetation | MovingCar | StaticCar | Human | mIoU |
|---|---|---|---|---|---|---|---|---|---|---|
| MSD (Lyu et al., 2020) | - | 57.0 | 79.8 | 74.0 | 74.5 | 55.9 | 62.9 | 32.1 | 19.7 | 57.0 |

| | | | | | | | | | |
|---|---|---|---|---|---|---|---|---|---|
| CANet (Yang et al., 2021a) | - | 66.0 | 86.6 | 62.1 | 79.3 | **78.1** | 47.8 | **68.3** | 19.9 | 63.5 |
| DANet (Fu et al., 2019) | ResNet18 | 64.9 | 85.9 | 77.9 | 78.3 | 61.5 | 59.6 | 47.4 | 9.1 | 60.6 |
| SwiftNet (Oršić and Šegvić, 2021) | ResNet18 | 64.1 | 85.3 | 61.5 | 78.3 | 76.4 | 51.1 | 62.1 | 15.7 | 61.1 |
| BiSeNet (Yu et al., 2018) | ResNet18 | 64.7 | 85.7 | 61.1 | 78.3 | 77.3 | 48.6 | 63.4 | 17.5 | 61.5 |
| MANet (Li et al., 2021b) | ResNet18 | 64.5 | 85.4 | 77.8 | 77.0 | 60.3 | 67.2 | 53.6 | 14.9 | 62.6 |
| ABCNet (Li et al., 2021c) | ResNet18 | 67.4 | 86.4 | 81.2 | 79.9 | 63.1 | 69.8 | 48.4 | 13.9 | 63.8 |
| Segmenter (Strudel et al., 2021) | ViT-Tiny | 64.2 | 84.4 | 79.8 | 76.1 | 57.6 | 59.2 | 34.5 | 14.2 | 58.7 |
| SegFormer (Xie et al., 2021) | MiT-B1 | 66.6 | 86.3 | 80.1 | 79.6 | 62.3 | 72.5 | 52.5 | 28.5 | 66.0 |
| BANet (Wang et al., 2021b) | ResT-Lite | 66.7 | 85.4 | 80.7 | 78.9 | 62.1 | 69.3 | 52.8 | 21.0 | 64.6 |
| BoTNet (Srinivas et al., 2021) | ResNet18 | 64.5 | 84.9 | 78.6 | 77.4 | 60.5 | 65.8 | 51.9 | 22.4 | 63.2 |
| CoaT (Xu et al., 2021) | CoaT-Mini | **69.0** | **88.5** | 80.0 | 79.3 | 62.0 | 70.0 | 59.1 | 18.9 | 65.8 |
| UNetFormer | ResNet18 | 68.4 | 87.4 | **81.5** | **80.2** | 63.5 | **73.6** | 56.4 | **31.0** | **67.8** |

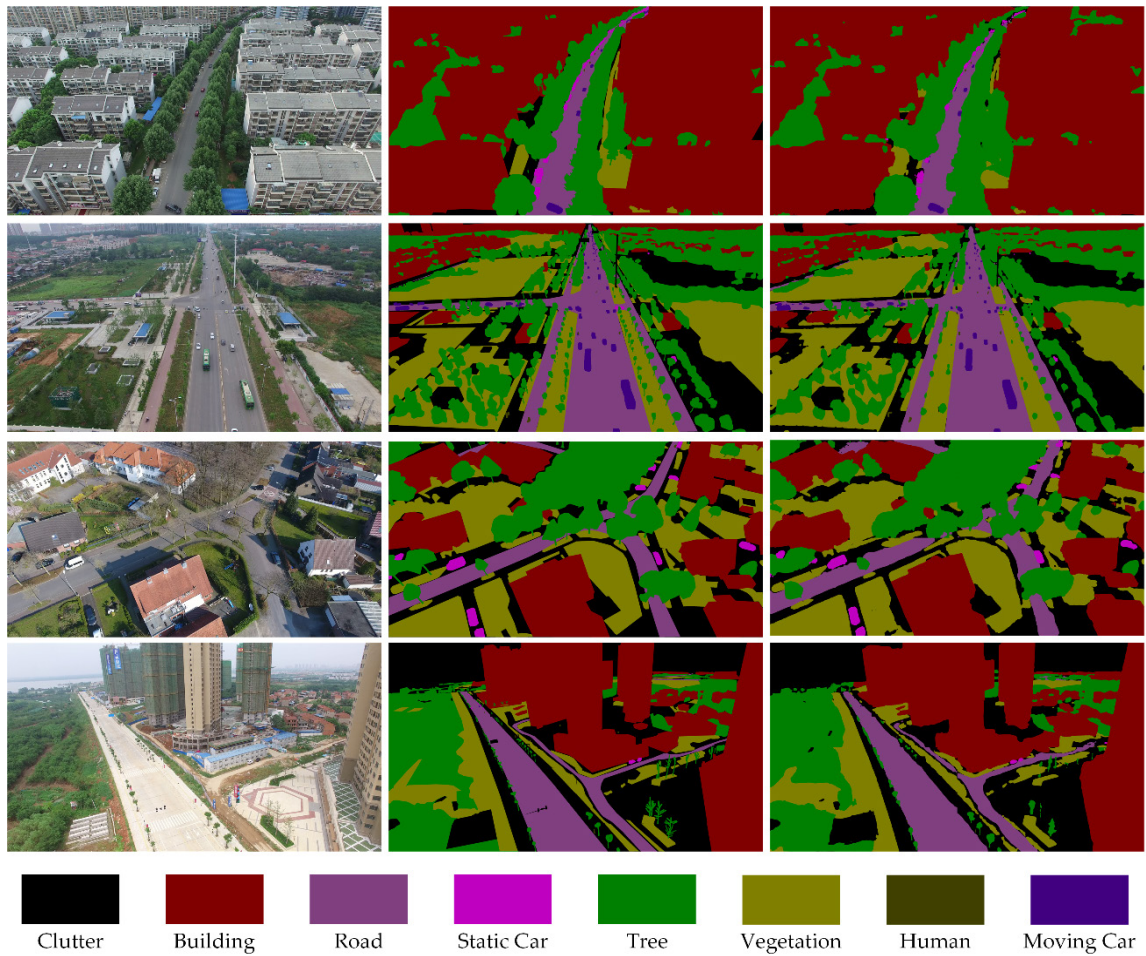

| | | | | | | | | |
|---|---|---|---|---|---|---|---|
| ⬛ Clutter | 🟥 Building | 🟪 Road | 🟪 Static Car | 🟩 Tree | 🟨 Vegetation | 🟫 Human | 🟪 Moving Car |

**Fig. 6.** Segmentation results from the UAVid validation set. The first column represents the input

RGB images. The second column denotes the ground reference. The third column shows the segmentation maps produced by our method.

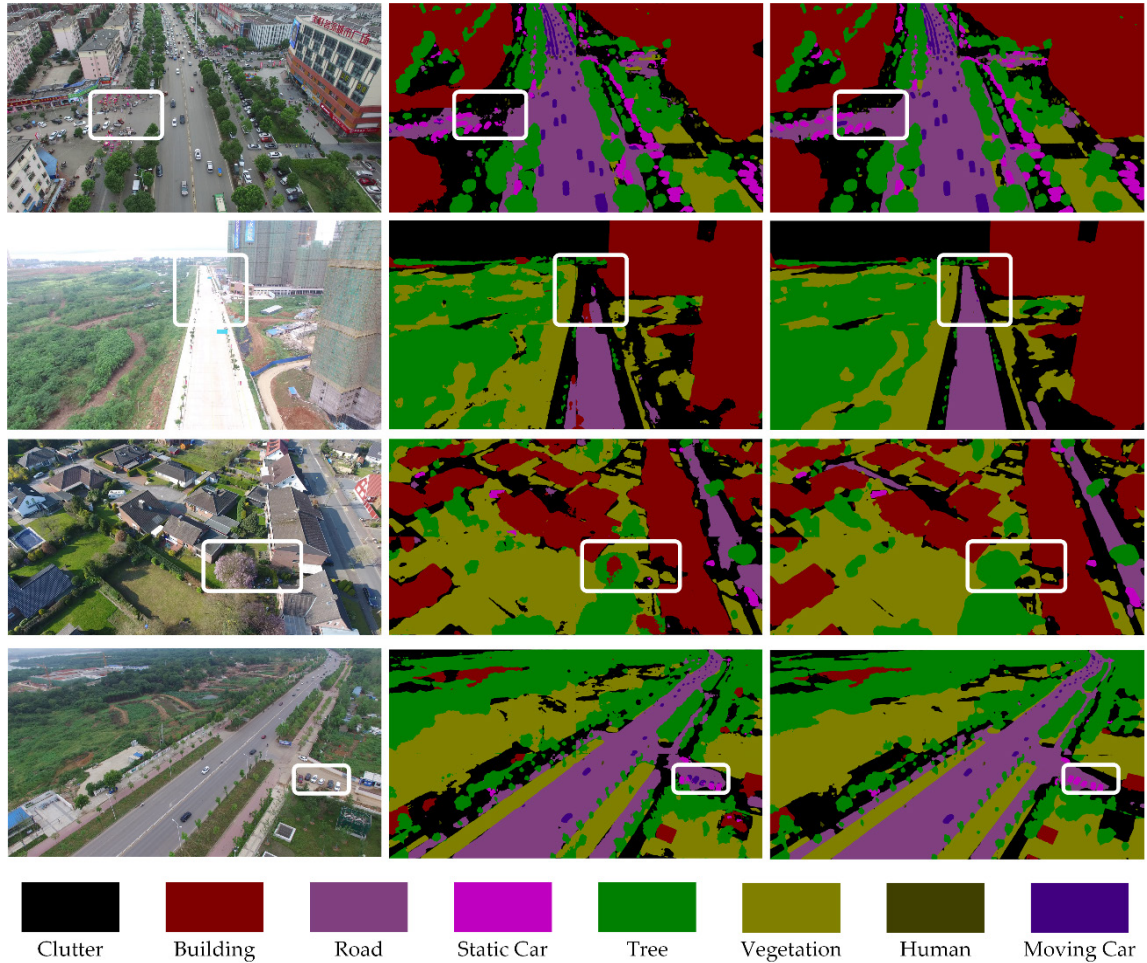

| Clutter | Building | Road | Static Car | Tree | Vegetation | Human | Moving Car |

**Fig. 7.** Enlarged visualization of results from the UAVid test set. The first column represents the input RGB images. The second column denotes the segmentation results of the baseline. The third column shows the segmentation maps of our method.

### 4.3.3 Results on the Vaihingen and Potsdam dataset

The ISPRS Vaihingen and Potsdam are two widely-used datasets for segmentation tasks. Numerically high accuracies have been achieved by the specially designed models on these two datasets. In this section, we demonstrate that our UNetFormer can not only surpass lightweight models but also obtain competitive scores in comparison with leading networks.

As illustrated in **Table 8**, the proposed UNetFormer delivers the best F1, OA and mIoU on the Vaihingen test set, outperforming other CNN-based and Transformer-based lightweight networks by a significant margin. It is worth noting that our method yields an 88.5% F1 score on the "car" class, exceeding other networks by more than 1.7%. Moreover, the prediction results of ID 2 and 22 are shown in **Fig. 8**, while the enlarged visualization of results is illustrated in **Fig. 9** (Top), which also demonstrates the effectiveness of our method.

**TABLE 8**. Quantitative comparison results on the Vaihingen test set with state-of-the-art lightweight networks. The best values in the column are in bold.

| Method | Backbone | Imp.surf. | Building | Lowveg. | Tree | Car | MeanF1 | OA | mIoU |
|---|---|---|---|---|---|---|---|---|---|
| DABNet (Li et al., 2019) | - | 87.8 | 88.8 | 74.3 | 84.9 | 60.2 | 79.2 | 84.3 | 70.2 |
| ERFNet (Romera et al., 2017) | - | 88.5 | 90.2 | 76.4 | 85.8 | 53.6 | 78.9 | 85.8 | 69.1 |
| BiSeNet (Yu et al., 2018) | ResNet18 | 89.1 | 91.3 | 80.9 | 86.9 | 73.1 | 84.3 | 87.1 | 75.8 |
| PSPNet (Zhao et al., 2017a) | ResNet18 | 89.0 | 93.2 | 81.5 | 87.7 | 43.9 | 79.0 | 87.7 | 68.6 |
| DANet (Fu et al., 2019) | ResNet18 | 90.0 | 93.9 | 82.2 | 87.3 | 44.5 | 79.6 | 88.2 | 69.4 |
| FANet (Hu et al., 2020) | ResNet18 | 90.7 | 93.8 | 82.6 | 88.6 | 71.6 | 85.4 | 88.9 | 75.6 |
| EaNet (Zheng et al., 2020a) | ResNet18 | 91.7 | 94.5 | 83.1 | 89.2 | 80.0 | 87.7 | 89.7 | 78.7 |
| ShelfNet (Zhuang et al., 2019) | ResNet18 | 91.8 | 94.6 | 83.8 | 89.3 | 77.9 | 87.5 | 89.8 | 78.3 |
| MAResU-Net (Li et al., 2021a) | ResNet18 | 92.0 | 95.0 | 83.7 | 89.3 | 78.3 | 87.7 | 90.1 | 78.6 |
| SwiftNet (Oršić and Šegvić, 2021) | ResNet18 | 92.2 | 94.8 | 84.1 | 89.3 | 81.2 | 88.3 | 90.2 | 79.6 |
| ABCNet (Li et al., 2021c) | ResNet18 | **92.7** | 95.2 | 84.5 | 89.7 | 85.3 | 89.5 | 90.7 | 81.3 |
| BoTNet (Srinivas et al., 2021) | ResNet18 | 89.9 | 92.1 | 81.8 | 88.7 | 71.3 | 84.8 | 88.0 | 74.3 |
| BANet (Wang et al., 2021b) | ResT-Lite | 92.2 | 95.2 | 83.8 | 89.9 | 86.8 | 89.6 | 90.5 | 81.4 |
| Segmenter (Strudel et al., 2021) | ViT-Tiny | 89.8 | 93.0 | 81.2 | 88.9 | 67.6 | 84.1 | 88.1 | 73.6 |
| UNetFormer | ResNet18 | **92.7** | **95.3** | **84.9** | **90.6** | **88.5** | **90.4** | **91.0** | **82.7** |

For a comprehensive evaluation, we further conducted experiments on the Postdam dataset.

As shown in **Table 10**, our UNetFormer achieves a 92.8% mean F1 score and an 86.8% mIoU on the Potsdam test set. The results of the UNetFormer not only exceed the excellent convolutional lightweight network ABCNet (Li et al., 2021c) but also outperform recent Transformer-based lightweight networks, such as Segmenter (Strudel et al., 2021) and BANet (Wang et al., 2021b). We also provide segmentation results for ID 3_14 and 2_13 (**Fig. 9**) and an enlarged visualization of the results (**Fig. 10** Bottom) to show the preferential performance of our network.

**TABLE 9** Quantitative comparison results on the Potsdam test set with state-of-the-art lightweight networks. The best values in the column are in bold.

| Method | Backbone | Imp.surf. | Building | Lowveg. | Tree | Car | MeanF1 | OA | mIoU |
|---|---|---|---|---|---|---|---|---|---|
| ERFNet (Romera et al., 2017) | - | 88.7 | 93.0 | 81.1 | 75.8 | 90.5 | 85.8 | 84.5 | 76.2 |
| DABNet (Li et al., 2019) | - | 89.9 | 93.2 | 83.6 | 82.3 | 92.6 | 88.3 | 86.7 | 79.6 |
| BiSeNet (Yu et al., 2018) | ResNet18 | 90.2 | 94.6 | 85.5 | 86.2 | 92.7 | 89.8 | 88.2 | 81.7 |
| EaNet (Zheng et al., 2020a) | ResNet18 | 92.0 | 95.7 | 84.3 | 85.7 | 95.1 | 90.6 | 88.7 | 83.4 |
| MAResU-Net (Li et al., 2021a) | ResNet18 | 91.4 | 95.6 | 85.8 | 86.6 | 93.3 | 90.5 | 89.0 | 83.9 |
| DANet (Fu et al., 2019) | ResNet18 | 91.0 | 95.6 | 86.1 | 87.6 | 84.3 | 88.9 | 89.1 | 80.3 |
| SwiftNet (Oršić and Šegvić, 2021) | ResNet18 | 91.8 | 95.9 | 85.7 | 86.8 | 94.5 | 91.0 | 89.3 | 83.8 |
| FANet (Hu et al., 2020) | ResNet18 | 92.0 | 96.1 | 86.0 | 87.8 | 94.5 | 91.3 | 89.8 | 84.2 |
| ShelfNet (Zhuang et al., 2019) | ResNet18 | 92.5 | 95.8 | 86.6 | 87.1 | 94.6 | 91.3 | 89.9 | 84.4 |
| ABCNet (Li et al., 2021c) | ResNet18 | 93.5 | 96.9 | **87.9** | **89.1** | 95.8 | 92.7 | **91.3** | 86.5 |
| Segmenter (Strudel et al., 2021) | ViT-Tiny | 91.5 | 95.3 | 85.4 | 85.0 | 88.5 | 89.2 | 88.7 | 80.7 |
| BANet (Wang et al., 2021b) | ResT-Lite | 93.3 | 96.7 | 87.4 | **89.1** | 96.0 | 92.5 | 91.0 | 86.3 |
| SwinUperNet (Liu et al., 2021) | Swin-Tiny | 93.2 | 96.4 | 87.6 | 88.6 | 95.4 | 92.2 | 90.9 | 85.8 |
| UNetFormer | ResNet18 | **93.6** | **97.2** | 87.7 | 88.9 | **96.5** | **92.8** | **91.3** | **86.8** |

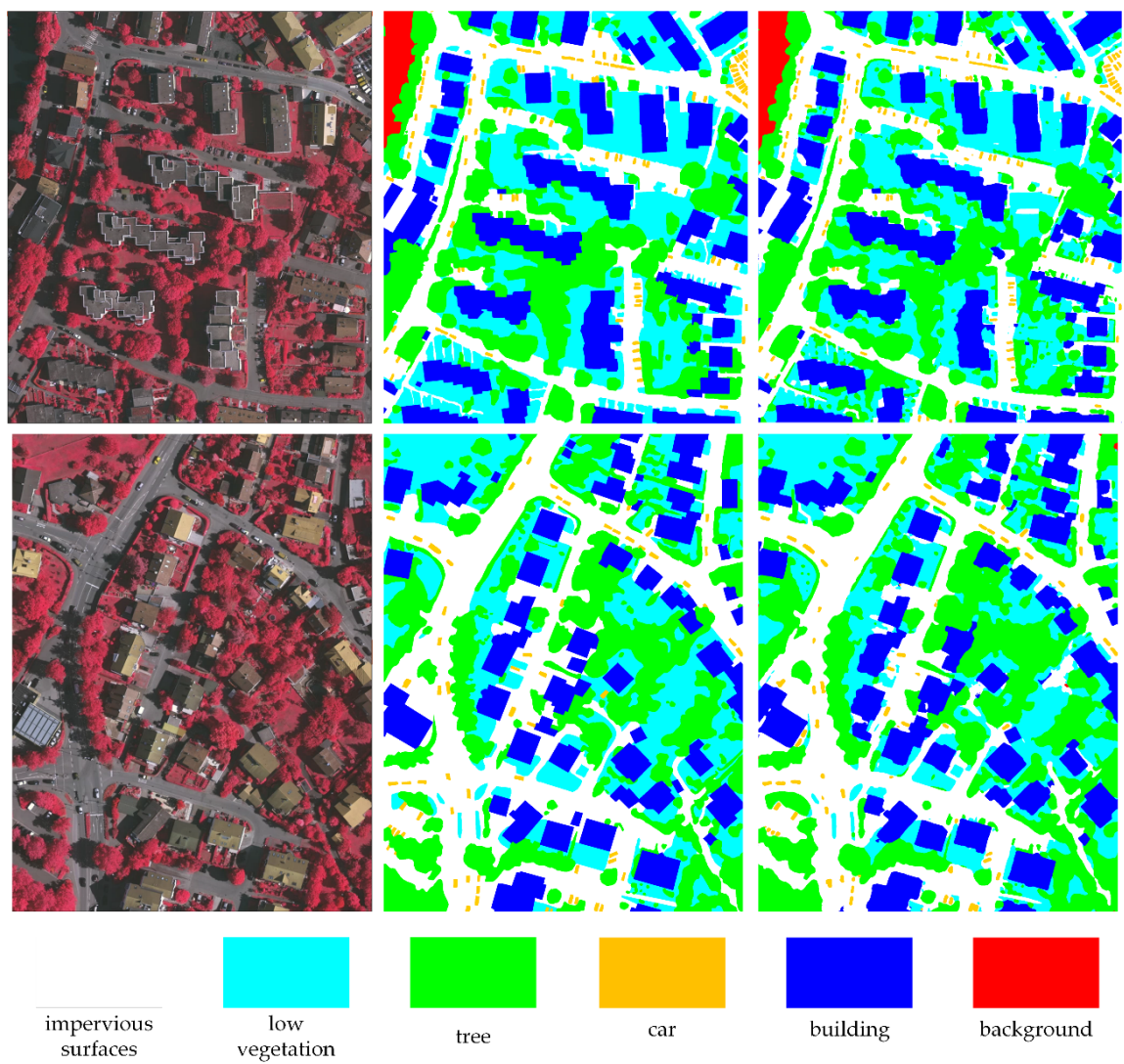

impervious surfaces | low vegetation | tree | car | building | background

**Fig. 8.** Visualization results of ID 2 and 22 from the Vaihingen test set. The first column denotes the input RGB images. The second column represents the ground truth. The third column shows the segmentation results of the proposed UNetFormer.

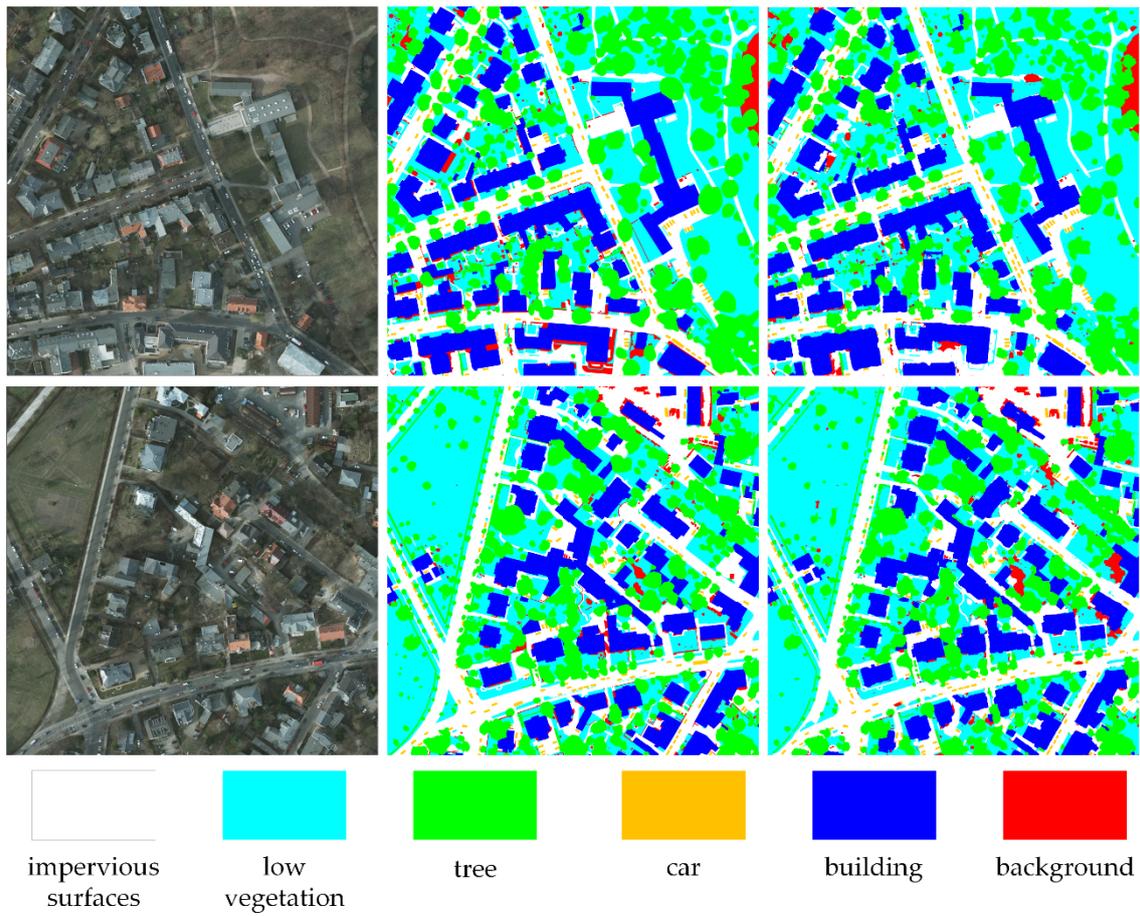

| | impervious surfaces | low vegetation | tree | car | building | background |
|---|---|---|---|---|---|---|

**Fig. 9.** Visualization results of ID 3_14 and 2_13 from the Potsdam test set. The first column denotes the input RGB images. The second column represents the ground truth. The third column shows the segmentation results of the proposed UNetFormer.

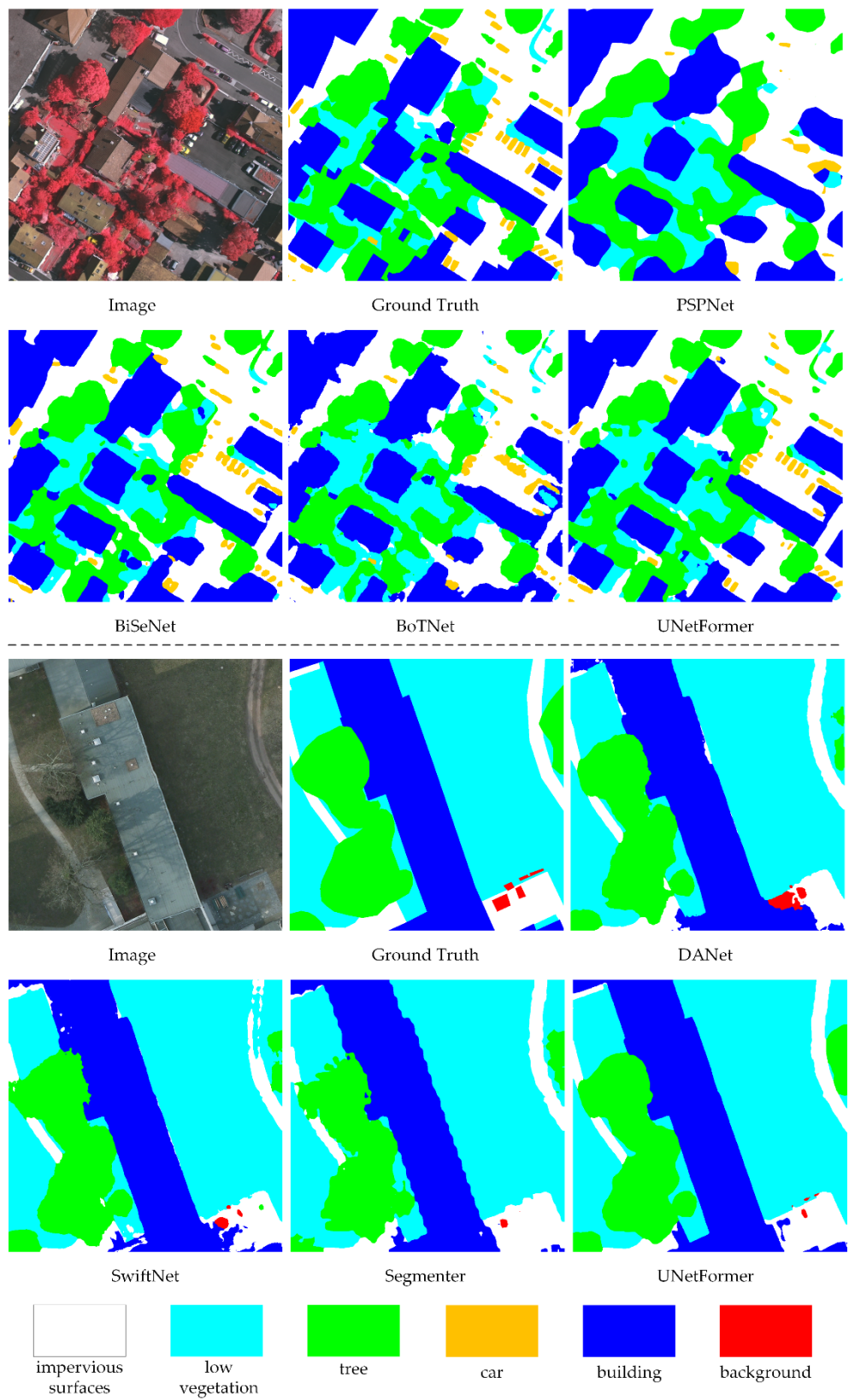

**Fig. 10.** Enlarged visualization of results from the Vaihingen (top) and Potsdam (bottom) test set.

**4.3.4 Results on the LoveDA dataset**

We undertook experiments on the LoveDA dataset to further evaluate the performance of the UNetFormer. Benefiting from the captured global-local context, the UNetFormer can handle both urban and rural scenes well in the LoveDA dataset. The comparison results are listed in **TABLE 10**. Remarkably, the UNetFormer obtains the highest mIoU (52.4%) with the least complexity and the fastest speed. Visualized comparisons are exhibited in **Fig. 11**.

**TABLE 10**. Quantitative comparison results on the LoveDA test set with other networks. The complexity and speed are measured by a 1024×1024 input on a single NVIDIA GTX 3090 GPU. The best values in the column are in bold.

| Method | Backbone | Background | Building | Road | Water. | Barren | Forest | Agriculture | mIoU | Complexity | Speed |
|---|---|---|---|---|---|---|---|---|---|---|---|
| PSPNet (Zhao et al., 2017a) | ResNet50 | 44.4 | 52.1 | 53.5 | 76.5 | 9.7 | 44.1 | 57.9 | 48.3 | 105.7 | 52.2 |
| DeepLabV3+ (Chen et al., 2018a) | ResNet50 | 43.0 | 50.9 | 52.0 | 74.4 | 10.4 | 44.2 | 58.5 | 47.6 | 95.8 | 53.7 |
| SemanticFPN (Kirillov et al., 2019) | ResNet50 | 42.9 | 51.5 | 53.4 | 74.7 | 11.2 | 44.6 | 58.7 | 48.2 | 103.3 | 52.7 |
| FarSeg (Zheng et al., 2020b) | ResNet50 | 43.1 | 51.5 | 53.9 | 76.6 | 9.8 | 43.3 | 58.9 | 48.2 | - | 47.8 |
| FactSeg (Ma et al., 2021) | ResNet50 | 42.6 | 53.6 | 52.8 | 76.9 | 16.2 | 42.9 | 57.5 | 48.9 | - | 46.7 |
| BANet (Wang et al., 2021b) | ResT-Lite | 43.7 | 51.5 | 51.1 | 76.9 | 16.6 | 44.9 | **62.5** | 49.6 | 52.6 | 11.5 |
| TransUNet (Chen et al., 2021b) | ViT-R50 | 43.0 | 56.1 | 53.7 | 78.0 | 9.3 | 44.9 | 56.9 | 48.9 | 803.4 | 13.4 |
| Segmenter (Strudel et al., 2021) | ViT-Tiny | 38.0 | 50.7 | 48.7 | 77.4 | 13.3 | 43.5 | 58.2 | 47.1 | 26.8 | 14.7 |
| SwinUperNet (Liu et al., 2021) | Swin-Tiny | 43.3 | 54.3 | 54.3 | 78.7 | 14.9 | 45.3 | 59.6 | 50.0 | 349.1 | 19.5 |
| DC-Swin (Wang et al., 2022) | Swin-Tiny | 41.3 | 54.5 | **56.2** | 78.1 | 14.5 | **47.2** | 62.4 | 50.6 | 183.8 | 23.6 |
| UNetFormer | ResNet18 | **44.7** | **58.8** | 54.9 | **79.6** | **20.1** | 46.0 | **62.5** | **52.4** | **46.9** | **115.3** |

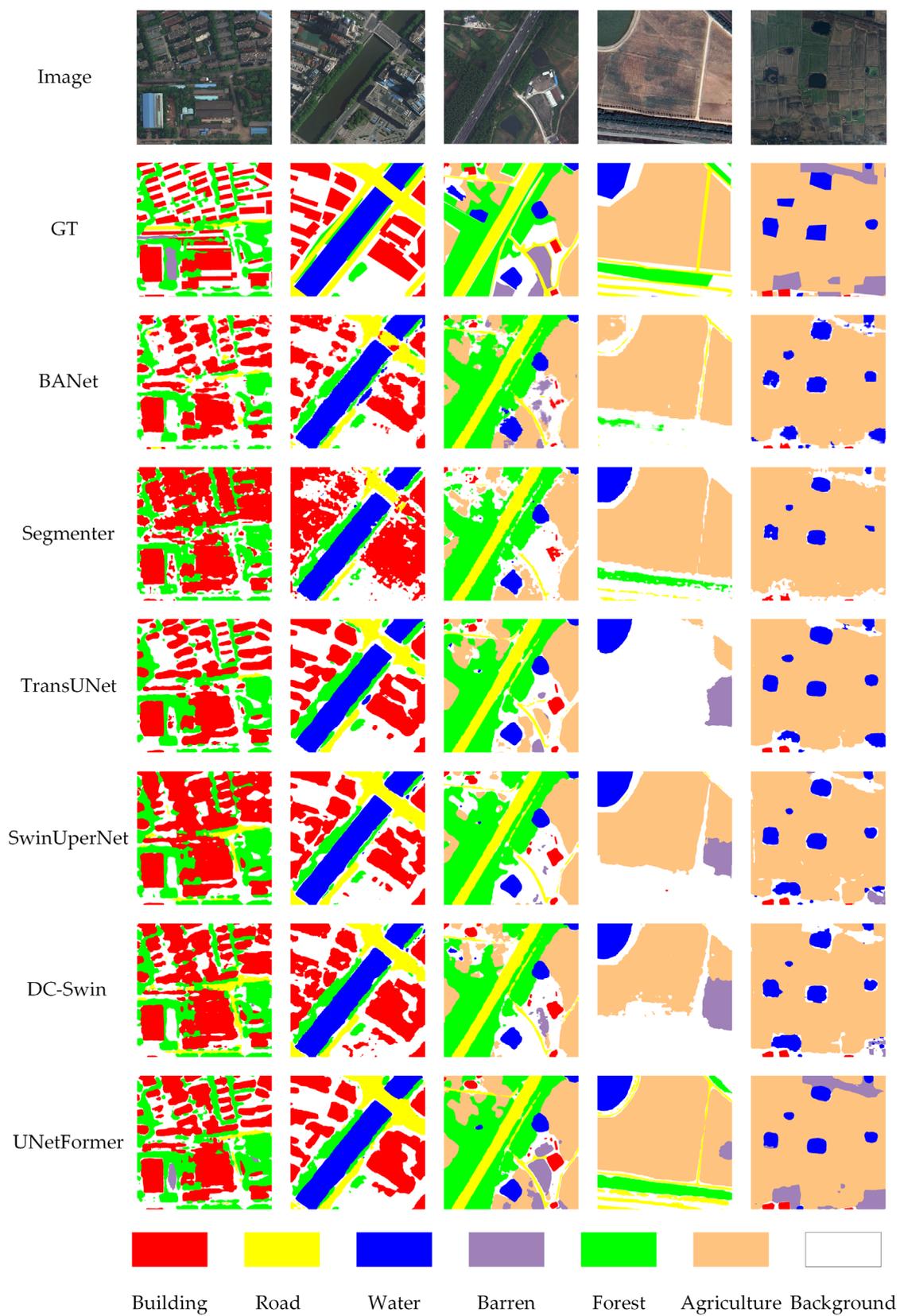

**Fig. 11.** Visualization comparisons on the LoveDA validation set.

## 5. Discussion

### 5.1 Global-local context

The advantage of the dual-branch structure of the proposed efficient global-local attention is that it can extract sufficient global contextual information while preserving fine-grained local information. To demonstrate this, we visualise the feature maps from the efficient global-local attention in **Fig. 12**. As can be seen, the local context extracted by the local branch preserves the abundant local features but lacks spatial consistency, while the global context captured by the global branch has a more consistent character but lacks locality. Meanwhile, for the global branch, performing the self-attention operation within a local window also causes jagged edges in the window context. We address this issue by employing a cross-shaped window context interaction module for context aggregation. By this means, the interaction between windows is enhanced, thereby resolving the jaggedness issue. Notably, the extracted global-local context with both locality and spatial consistency is visibly superior to the single global context or local context.

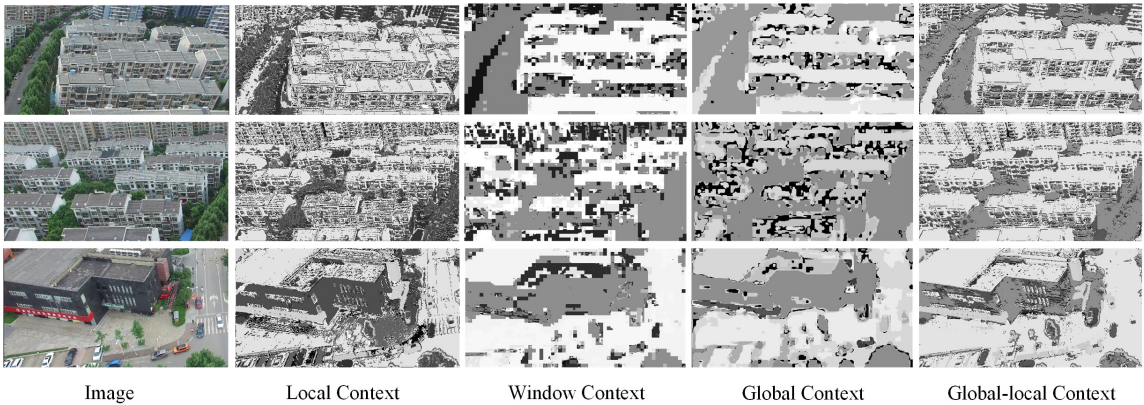

| Image | Local Context | Window Context | Global Context | Global-local Context |

**Fig. 12.** Visualization of the local context, window context, global context and global-local context in the proposed efficient global-local attention.

## 5.2 Model efficiency

The proposed UNetFormer adopts a hybrid structure with a CNN-based encoder and a Transformer-based decoder to achieve real-time performance. This hybrid design demonstrates superiority compared to other encoder-decoder combinations (**TABLE 5**). Moreover, the efficient global-local attention module utilizes the cross-shaped window context interaction module to replace the shift window attention for capturing cross-window relationships, which further increases the efficiency (**TABLE 2**). The superior trade-off between accuracy and efficiency brings advantages, such as the potential for the proposed UNetFormer to process real-time UAV images for environmental perception and monitoring in urban areas.

## 5.3 Transformer-based encoder

As shown in **TABLEs 4 and 5**, Transformers make strong encoders but greatly reduce the speed. Although Transformer-based encoders are not suitable for real-time applications, demonstrate advantages in pursuing high precision. Thus, we construct a fully Transformer-based network (FT-UNetFormer) to further explore the potential of the proposed Transformer-based decoder. To compare with state-of-the-art models at a similar level, we replace the lightweight ResNet18 encoder with he Swin Transformer (Swin-Base) (Liu et al., 2021). As listed in **TABLE 11**, the FT-UNetFormer yields the state-of-the-art results (91.3% F1 score and 84.1% mIoU) on the Vaihingen test set and outperforms other networks by at least 0.3% in F1 score. For the Potsdam dataset, our method also achieves competitive results (**TABLE 12**). These results further demonstrate the effectiveness of the proposed Transformer-based decoder and its potential in a fully Transformer structure.

**TABLE 11**. Quantitative comparison results on the Vaihingen test set with the state-of-the-art networks.

| Method | Backbone | Imp.surf. | Building | Lowveg. | Tree | Car | MeanF1 | OA | mIoU |
|---|---|---|---|---|---|---|---|---|---|
| CASIA2 (Liu et al., 2018) | ResNet101 | 93.2 | 96.0 | 84.7 | 89.9 | 86.7 | 90.1 | 91.1 | - |
| V-FuseNet (Audebert et al., 2018) | FuseNet | 91.0 | 94.4 | 84.5 | 89.9 | 86.3 | 89.2 | 90.0 | - |
| DLR_9 (Marmanis et al., 2018) | SegNet+VGG+FCN | 92.4 | 95.2 | 83.9 | 89.9 | 81.2 | 88.5 | 90.3 | - |
| RoteEqNet (Marcos et al., 2018) | - | 89.5 | 94.8 | 77.5 | 86.5 | 72.6 | 84.2 | 87.5 | - |
| UFMG_4 (Nogueira et al., 2019) | - | 91.1 | 94.5 | 82.9 | 88.0 | 81.3 | 87.7 | 89.4 | - |
| HUSTW5 (Sun et al., 2019) | SegNet | 93.3 | 96.1 | **86.4** | **90.8** | 74.6 | 88.2 | 91.6 | - |
| TreeUNet (Yue et al., 2019) | - | 92.5 | 94.9 | 83.6 | 89.6 | 85.9 | 89.3 | 90.4 | - |
| EaNet (Zheng et al., 2020a) | ResNet101 | 93.4 | 96.2 | 85.6 | 90.5 | 88.3 | 90.8 | 91.2 | - |
| DDCM-Net (Liu et al., 2020) | ResNet50 | 92.7 | 95.3 | 83.3 | 89.4 | 88.3 | 89.8 | 90.4 | - |
| MANet (Li et al., 2021b) | ResNe50 | 93.0 | 95.5 | 84.6 | 90.0 | 88.9 | 90.4 | 91.0 | 82.7 |
| AFNet (Yang et al., 2021b) | ResNet50+18 | 93.1 | **96.5** | 85.8 | 90.6 | 88.8 | 91.0 | **91.7** | - |
| HMANet (Niu et al., 2021) | ResNet101 | **93.5** | 95.9 | 85.4 | 90.4 | 89.6 | 91.0 | 91.4 | 83.5 |
| STransFuse (Gao et al., 2021) | - | 88.3 | 91.46 | 79.0 | 85.5 | 77.1 | 78.7 | 86.1 | 66.7 |
| BoTNet (Srinivas et al., 2021) | ResNet50 | 92.2 | 95.3 | 83.9 | 90.0 | 85.5 | 89.4 | 90.5 | 81.1 |
| SwinUperNet (Liu et al., 2021) | Swin-Small | 92.8 | 95.6 | 85.1 | 90.6 | 85.1 | 89.8 | 91.0 | 81.8 |
| SwinB-CNN+BD (Zhang et al., 2022) | Swin-Base | 95.3 | 86.9 | 83.6 | 92.2 | 89.6 | 89.5 | 90.4 | - |
| DC-Swin (Wang et al., 2022) | Swin-Small | 93.6 | 96.2 | 85.8 | 90.4 | 87.6 | 90.7 | 91.6 | 83.2 |
| FT-UNetFormer | Swin-Base | **93.5** | 96.0 | 85.6 | **90.8** | **90.4** | **91.3** | 91.6 | **84.1** |

**TABLE 12** Quantitative comparison results on the Potsdam test set with state-of-the-art networks.

| Method | Backbone | Imp.surf. | Building | Low. veg. | Tree | Car | MeanF1 | OA | mIoU |
|---|---|---|---|---|---|---|---|---|---|
| DST_5 (Sherrah, 2016) | FCN | 92.5 | 96.4 | 86.7 | 88.0 | 94.7 | 91.7 | 90.3 | - |
| V-FuseNet (Audebert et al., 2018) | FuseNet | 92.7 | 96.3 | 87.3 | 88.5 | 95.4 | 92.0 | 90.6 | - |
| SWJ_2 | ResNet101 | **94.4** | 97.4 | 87.8 | 87.6 | 94.7 | 92.4 | 91.7 | - |
| AMA_1 | - | 93.4 | 96.8 | 87.7 | 88.8 | 96.0 | 92.5 | 91.2 | - |
| UFMG_4 (Nogueira et al., 2019) | - | 90.8 | 95.6 | 84.4 | 84.3 | 92.4 | 89.5 | 87.9 | - |
| S-RA-FCN (Mou et al., 2020) | VGG16 | 91.3 | 94.7 | 86.8 | 83.5 | 94.5 | 90.2 | 88.6 | 82.4 |
| HUSTW4 (Sun et al., 2019) | ResegNets | 93.6 | **97.6** | 88.5 | 88.8 | 94.6 | 92.6 | 91.6 | - |
| TreeUNet (Yue et al., 2019) | - | 93.1 | 97.3 | 86.8 | 87.1 | 95.8 | 92.0 | 90.7 | - |
| ResUNet-a (Diakogiannis et al., 2020) | - | 93.5 | 97.2 | 88.2 | 89.2 | 96.4 | 92.9 | 91.5 | - |

| | | | | | | | | |
|---|---|---|---|---|---|---|---|---|
| DDCM-Net (Liu et al., 2020) | ResNet50 | 92.9 | 96.9 | 87.7 | 89.4 | 94.9 | 92.3 | 90.8 | - |
| LANet (Ding et al., 2021) | ResNet50 | 93.1 | 97.2 | 87.3 | 88.0 | 94.2 | 92.0 | 90.8 | - |
| AFNet (Yang et al., 2021b) | ResNet50+18 | 94.1 | **97.6** | 88.7 | 89.7 | **97.1** | **93.4** | 92.1 | - |
| HMANet (Niu et al., 2021) | ResNet101 | 93.9 | **97.6** | 88.7 | 89.1 | 96.8 | 93.2 | **92.2** | 87.3 |
| STransFuse (Gao et al., 2021) | - | 89.8 | 93.9 | 82.9 | 83.6 | 88.5 | 82.1 | 86.7 | 71.5 |
| SwinB-CNN+BD (Zhang et al., 2022) | Swin-Base | 92.2 | 95.3 | 83.6 | 89.2 | 86.9 | 89.4 | 90.4 | - |
| SwinTF-FPN (Panboonyuen et al., 2021) | Swin-Small | 93.3 | 96.8 | 87.8 | 88.8 | 95.0 | 92.3 | 91.1 | 85.9 |
| ResT (Zhang and Yang, 2021) | ResT-Base | 92.7 | 96.1 | 87.5 | 88.6 | 94.8 | 91.9 | 90.6 | 85.2 |
| FT-UNetFormer | Swin-Base | 93.9 | 97.2 | **88.8** | **89.8** | 96.6 | 93.3 | 92.0 | **87.5** |

## 6. Conclusion

In this paper, we proposed a novel Transformer-based decoder and constructed a UNet-like Transformer (UNetFormer) for efficient semantic segmentation of remotely sensed urban scene images. Since global and local contexts are both crucial for urban scene segmentation, we designed a global-local Transformer block (GLTB) to construct the decoder and developed a feature refinement head (FRH) to optimize the extracted global-local context. For efficient segmentation, the proposed Transformer-based decoder was combined with a lightweight CNN-based encoder. A comprehensive set of benchmark experiments and ablation studies on the ISPRS Vaihingen and Potsdam datasets and the UAVid dataset as well as the LoveDA dataset demonstrated the effectiveness and efficiency of the proposed method for real-time urban applications. Furthermore, the proposed Transformer-based decoder also works well in a fully Transformer structure and obtains state-of-the-art performance on the Vaihingen dataset. In future research, we will continue to explore the potential and feasibility of the Transformer for geospatial vision tasks.

## Declaration of Competing Interest

The authors declare that they have no known competing financial interests or personal relationships that could have appeared to influence the work reported in this paper.